\documentclass[10pt,twocolumn,letterpaper]{article}

\usepackage{iccv}
\usepackage{amsfonts} 
\usepackage[accsupp]{axessibility}
\usepackage{dblfloatfix}
\usepackage{nicefrac}       
\usepackage{microtype}      
\usepackage{graphicx}
\usepackage{amsmath}
\usepackage{amssymb}
\usepackage{pifont}
\usepackage{booktabs}
\usepackage{times}
\usepackage{graphicx}
\usepackage{amsmath}
\usepackage{amssymb}
\usepackage{dsfont}
\usepackage{siunitx}
\usepackage{multirow}
\usepackage{subcaption}
\usepackage{adjustbox}
\usepackage{enumitem}
\usepackage{tablefootnote}
\usepackage{bbm}
\usepackage{xspace}
\usepackage{mathtools}
\usepackage{xcolor}
\usepackage{color, colortbl}
\definecolor{Gray}{gray}{0.9}
\usepackage[normalem]{ulem}
\usepackage{tabularx}
\usepackage{array}
\usepackage{wrapfig}
\usepackage{flushend}
\usepackage{longtable}
\usepackage{outlines}
\usepackage[symbol]{footmisc}

\usepackage[accsupp]{axessibility}  

\definecolor{salmon}{rgb}{1.0, 0.55, 0.41}

\definecolor{citecolor}{HTML}{0071bc}
\usepackage[pagebackref,breaklinks,colorlinks,citecolor=citecolor]{hyperref}

\usepackage[capitalize]{cleveref}

\crefname{section}{Section}{Sections}
\Crefname{section}{Section}{Sections}
\Crefname{table}{Table}{Tables}
\crefname{table}{Table}{Tables}
\crefname{figure}{Figure}{Figures}
\Crefname{figure}{Figure}{Figures}

\definecolor{darkgreen}{rgb}{0.0,0.6,0.0}

\makeatletter
\newcommand{\thickhline}{%
    \noalign {\ifnum 0=`}\fi \hrule height 1pt
    \futurelet \reserved@a \@xhline
}

\newcommand{\ignorebig}[1]{}

\newcommand{\cmark}{\textcolor{darkgreen}{\ding{51}}}
\newcommand{\xmark}{\textcolor{red}{\ding{55}}}

\newcommand{\comment}[1]{}

\newcommand{\data}{HoloAssist\xspace}

\newcommand\minisection[1]{\vspace{1mm}\noindent \textbf{#1}}

\newcommand\footnoteref[1]{\protected@xdef\@thefnmark{\ref{#1}}\@footnotemark}

\usepackage[pagebackref,breaklinks,colorlinks]{hyperref}
\usepackage[capitalize]{cleveref}
\crefname{section}{Sec.}{Secs.}
\Crefname{section}{Section}{Sections}
\Crefname{table}{Table}{Tables}
\crefname{table}{Tab.}{Tabs.}

\iccvfinalcopy 

\ificcvfinal\fi

\begin{document}

\title{\data: an Egocentric Human Interaction Dataset \\ for Interactive AI Assistants in the Real World}
\author{Xin Wang$^{1*}$~~ Taein Kwon$^{1,2*}$~~ Mahdi Rad$^{1}$~~ Bowen Pan$^{1\dagger}$~~ Ishani Chakraborty$^{1}$~~ Sean Andrist$^{1}$ \\
~~ Dan Bohus$^{1}$~~ Ashley Feniello$^{1}$~~ Bugra Tekin$^{1\dagger}$~~ Felipe Vieira Frujeri$^{1}$~~ Neel Joshi$^{1}$~~ Marc Pollefeys$^{1,2}$
\\ $^{1}$Microsoft~~~~ $^{2}$ ETH Zurich \\
}
\maketitle

\footnotetext[0]{*Co-first authors; $\dagger$Work done at Microsoft}

\begin{abstract}
Building an interactive AI assistant that can perceive, reason, and collaborate with humans in the real world has been a long-standing pursuit in the AI community. This work is part of a broader research effort to develop intelligent agents that can interactively guide humans through performing tasks in the physical world.
 As a first step in this direction, 
we introduce \data, a large-scale 
egocentric human interaction dataset, where two people collaboratively complete physical manipulation tasks.
The task performer executes the task while wearing a mixed-reality headset
that captures seven synchronized data streams. The task instructor watches the performer’s 
egocentric video in real time and guides them verbally.
By augmenting the data with action and conversational annotations and observing the rich behaviors of various participants, we present key insights into how human assistants correct mistakes, intervene in the task completion procedure, and ground their 
instructions to the environment. \data spans 166 hours of data captured by 350 
unique instructor-performer pairs. Furthermore, we construct and present benchmarks on mistake detection, intervention type prediction, and hand forecasting, along with detailed analysis. We expect \data will provide an important resource for building AI assistants that can fluidly collaborate with humans in the real world. Data can be downloaded at \small{\url{https://holoassist.github.io/}}. 
\end{abstract}

\section{Introduction}
\label{sec:intro}

\begin{figure}[t]
    \centering
    \includegraphics[width=\linewidth]{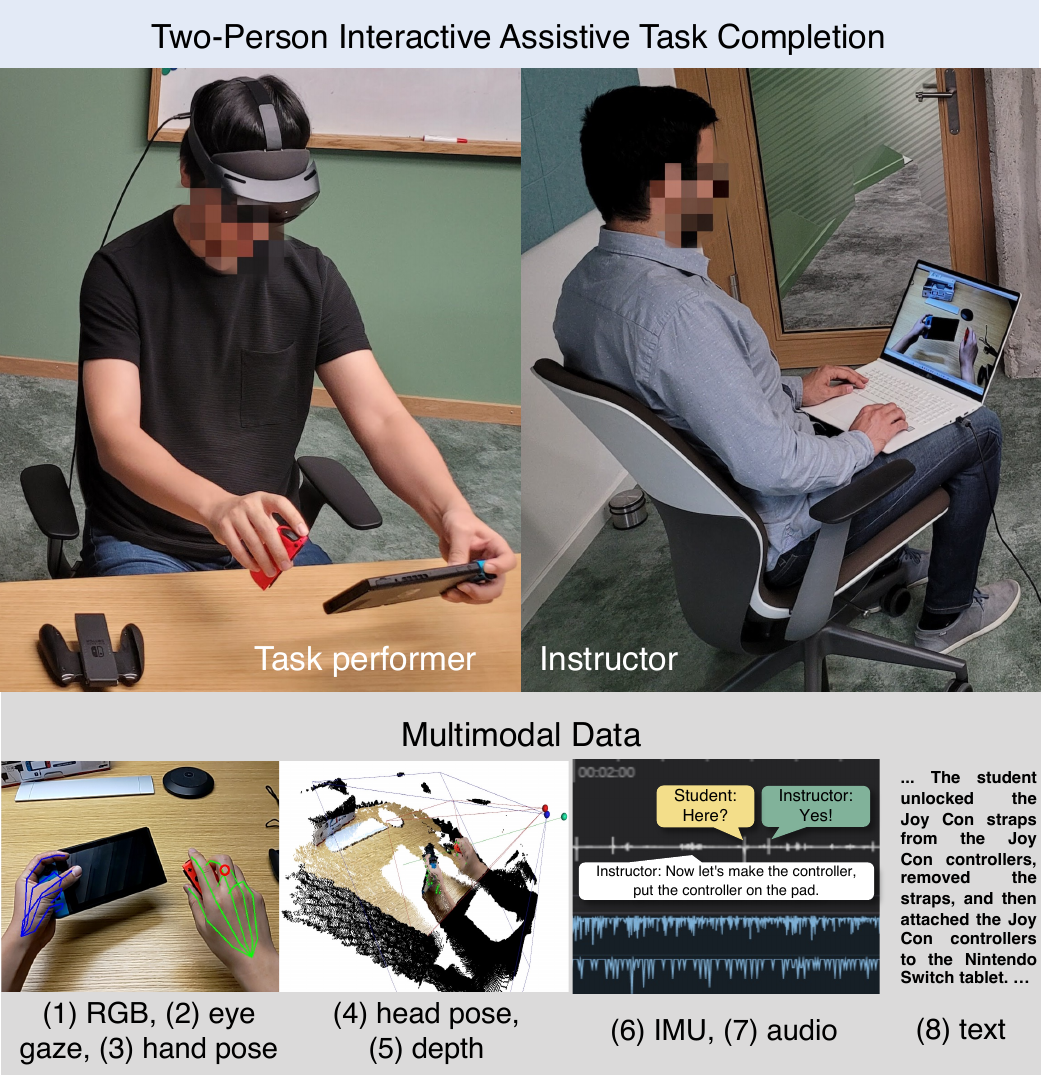}
    \caption{\data features a two-person interactive assistive task completion setting. The task performer wears an AR device and completes the
    tasks while the captured data is streamed over the network to a remote instructor watching it on the laptop. The instructor provides verbal guidance to the student. \data includes seven modalities captured live and human annotated text descriptions as the 8th modality.} \vspace{-4mm}
    \label{fig:teaser}
\end{figure}

Recent years have witnessed incredible progress in general-purpose AI agents that assist humans with various open-world tasks, especially in 
the digital world.  AI systems powered by large language models (LLMs) like ChatGPT~\cite{radford2019language} can answer users' questions and assist them with various 
text-based tasks. 
However, these AI assistants do not have sufficient first-hand experience in the physical world and thus cannot perceive world states and actively intervene in the task completion procedure. 

Building an AI assistant that can perceive, reason and interact in the physical world has attracted attention 
from researchers across different fields in computer vision~\cite{video2016, tldw, vislangnav, wong2022assistq}, human-computer interaction~\cite{aist2003talking,andrist2017looking,johnson1997steve,rich2007diamondhelp}, robotics~\cite{ahn2022can,shridhar2022perceiver}, and 
industrial practitioners. For example, AR Guides~\cite{guides}, which aims to guide users to complete complex tasks, has become popular with the 
development of augmented reality (AR) devices. However, existing systems often rely on pre-defined instructions or formulate the virtual assistant as a question 
answering~\cite{vislangnav, wong2022assistq} or video understanding problem~\cite{video2016, tldw} without real-world interaction. 

In another line of work, researchers have developed 
simulation environments like Habitat~\cite{habitat19iccv,szot2021habitat}, VirtualHome~\cite{puig2018virtualhome}, and AI2-Thor~\cite{kolve2017ai2} to build AI agents
that can interact with the physical world and 
collaboratively achieve new tasks~\cite{zholus2022iglu}. Still, a large gap remains in transferring these agents to the real world, and the interaction between agents is largely simplified compared to real-world human interaction. 

In this work, 
we focus on the challenges of developing intelligent agents that share perspectives with humans and interactively guide human users through performing tasks in the physical world. 
As a first step,  we introduce \data, a large-scale egocentric human interaction dataset to explore and identify the open problems in this direction.  As shown in Figure~\ref{fig:teaser}, the task 
\emph{performer} wears an AR headset\footnote{We use HoloLens 2~\cite{hololens} for data capture in this work.} to capture data while completing the tasks.  An 
\emph{instructor} watches the real-time egocentric video feed remotely and verbally guides 
the performer. We have developed and open-sourced a data capture tool~\cite{captureapp} using a distributed server-client setup to enable data streaming and multimodal data capture. 

\begin{figure}[t]
    \centering
    \includegraphics[width=1.\linewidth]{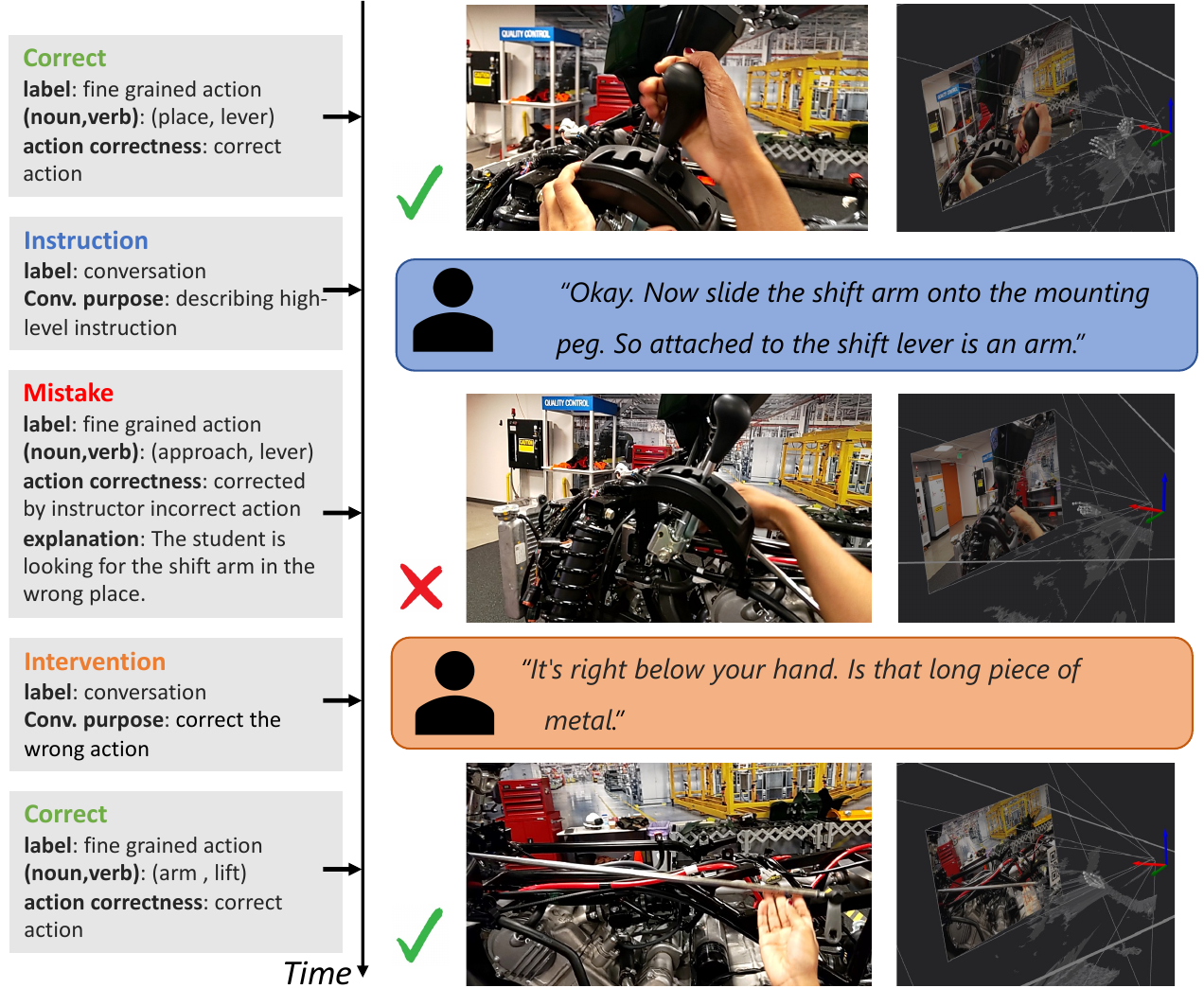}\vspace{-2mm}
    \caption{\data includes action and conversational annotations, in addition to text summaries of the videos, to indicate the mistakes and interventions in task completion. \emph{mistake} or \emph{correct} attributes are associated with each fine-grained action. A purpose label is associated with every utterance to indicate the type of verbal intervention. 
    }
    \label{fig:mistake_detection_example}\vspace{-1em}
\end{figure}

\data contains 166 hours of data captured by 222 diverse participants forming 350 unique instructor-performer pairs and carrying out
20 object-centric manipulation tasks. The objects range from 
common electronic devices to rare objects in factories and 
specialized labs. The tasks are generally challenging for first-time 
participants, requiring instructor assistance for successful completion. 
Seven raw sensor modalities are captured, including RGB, depth, head pose, 3D hand pose, eye gaze, audio, and IMU, to aid in the understanding of human intentions, estimating world states, predicting future actions, and so on.
Finally, the dataset is augmented with third-person manual annotations consisting of a text summary, intervention types, mistake annotation, and action segments of the videos as illustrated in Figure~\ref{fig:mistake_detection_example}. 

We have observed several characteristics
demonstrated by human instructors from \data. First, instructors  
are often proactive with precisely timed interventions. Instead 
of waiting until mistakes happen, instructors provide follow-up instructions when the task performer appears
confused. Second, the verbal guidance from the instructors 
tends to be concise and grounded in the task performer's 
environment. The instructions are often framed as spatial deictics to aid the task performer in spatial directions and distances in the 3D world.  Moreover, instructors often have a good world model estimation and can detect whether mistakes disrupt task completion and then adjust the guidance. 

 We take a step further and 
 introduce new tasks and benchmarks on mistake detection, intervention type prediction, and 3D hand pose forecasting, which we conjecture are  
 essential modules for an intelligent assistant. 
 Additionally, we benchmark the dataset on action classification and anticipation tasks and provide empirical results to understand the role of different modalities in various tasks. We hope our dataset, findings, and tooling can inspire and provide rich resources for future work on designing interactive AI assistants and situated AI assistance applications in the real world.

\section{Related Work}
\label{sec:related_work}
\vspace{-0.5em}
Our work closely connects with several lines of work in computer vision, especially egocentric vision, embodied AI, and human-computer interaction. 

\minisection{Interactive AI assistants.} Building interactive agents that can assist humans to carry out tasks in the world---real or virtual---has been a long-standing problem in 
different areas of AI and HCI~\cite{aist2003talking, bohus2005larri, johnson1997steve, manuvinakurike2022human, padmakumar2022teach, rich2007diamondhelp, rich2001collagen}. As far back as 1997, Johnson 
and Rickel introduced ``Steve''~\cite{johnson1997steve}, an 
early pedagogical agent that 
aims to help students learn procedural tasks in VR. Recent efforts have focused on new modeling approaches and data collection techniques for training conversational task guidance assistants, such as model-in-the-loop wizard-of-oz \cite{manuvinakurike2022human} and human-human interaction to mimic robot actions in simulated environments \cite{padmakumar2022teach}. In this work, we revisit this problem and provide a 
systematic study of real-world human interaction, and we also provide rich sensor information to push the frontiers of the research. 

\minisection{Egocentric video datasets.} Egocentric 
perspectives often convey rich information about the users' intentions. A shared perspective between the 
users and the human or AI assistants is useful for the assistants to provide more timely and grounded guidance. 
In computer vision, several egocentric video datasets~\cite{epickitchen, ego4d, kwon2021h2o, hoi4d, ragusa2021meccano, assembly101, egobody} have emerged in the community. EPIC-KICHENS~\cite{epickitchen} is a widely adopted egocentric video dataset capturing 
kitchen activities. The recent Ego4D dataset~\cite{ego4d} is the largest egocentric video dataset in the wild that provides a comprehensive database for egocentric perception in the 3D world. In contrast to earlier egocentric video datasets, \data features a multi-person interactive task completion setting, where human interaction during the procedure provides a rich source for designing AI assistants to be more proactive and grounded in the environment. 
Yet our work can benefit from the rich knowledge and representation learned from existing datasets like Ego4D and is complementary in nature.

\begin{figure}[t]
    \centering
    \includegraphics[width=.9\linewidth]{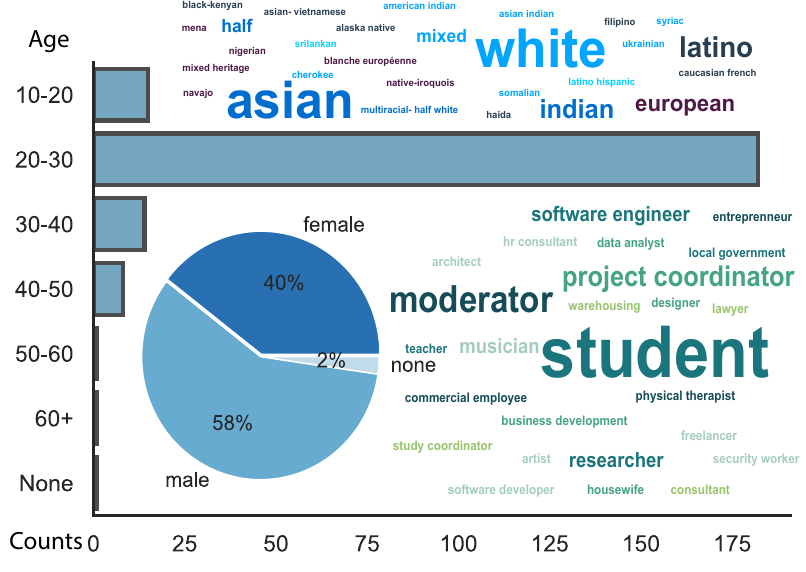}\vspace{-2mm}
    \caption{\data was collected by participants diverse in ages, occupations, genders, and geography. This helps us to study a diverse set of users with different backgrounds.}
    \label{fig:demo_graphics}\vspace{-1em}
\end{figure}

\minisection{Mistake detection.} One of the key observations in human interaction is that human assistants tend to correct mistakes and proactively intervene in the task completion procedure.  While there has been a large body of work for video-based anomaly detection~\cite{nayak2021comprehensive, anomaly2011,zhou2019anomalynet,opensetvideoanomaly}, mistake detection in procedural settings has been under-explored. The Assembly101 dataset~\cite{assembly101} proposes a mistake detection task to predict if a coarse-grained action segment is a mistake or correction. By contrast, \data emphasizes fine-grained actions since instructors may intervene when they spot a student's mistake in an active intervention setting rather than wait until the whole step (\ie, coarse-grained action) is completed. In addition, we propose a new intervention prediction task and, in combination, enable a more comprehensive understanding of interactions in an assistive task completion setting. 

\minisection{Multimodality and interaction.} Human 
interaction with the world is multimodal as we see, 
speak, and touch objects in the environment. In \data, we
collect seven raw sensor modalities that might help 
understand humans' intentions, estimate the world states, 
predict future actions, etc.  Previous datasets~\cite{epickitchen, ego4d, kwon2021h2o, egobody}
often provide a limited subset of modalities.  Although not every sensor may currently be relevant for the downstream tasks, 
the seven synchronized sensor modalities provided in \data
will give practitioners more potential for designing multimodal agents and models even beyond the scope of this work. 

\minisection{Embodied simulation platforms.} There is an emerging interest in embodied agents that can perceive, reason, and act in the 3D world. Researchers~\cite{kolve2017ai2, habitat19iccv, puig2018virtualhome, srivastava2022behavior, szot2021habitat, zholus2022iglu} build various simulation environments to learn such embodied agents.  IGLU~\cite{zholus2022iglu} aims to build interactive agents that learn to solve a task while being provided with grounded natural language instructions in a collaborative environment based on Minecraft, a popular video game. \data complements this line of work by providing more realistic human interaction and real-world sensor perception. 

\vspace{-1mm}
\section{HoloAssist: Human Assistance Dataset}
\vspace{-1mm}
In this work, we introduce \data which features a two-person collaboration scenario and can be used to situate
AI assistance in the physical world. We will start by describing the data collection and statistics in Section~\ref{sec:data_collection_scenario} and  annotations in Section~\ref{sec:annotation}, before diving into the observations and benchmarks in the following sections. 

\begin{table}[t]
\centering
\setlength{\extrarowheight}{4pt}%
\adjustbox{width=\linewidth}{
\begin{tabular}{@{}lm{.85\linewidth}@{}}
\textsc{Object Scales}  & \textsc{Object Categories}\\
\midrule
Small        & GoPro, Nintendo Switch, DSLR                            \\
\cmidrule{2-2}
Medium       & Portable printer,  Computer, Nespresso machine    \\  
\cmidrule{2-2}
Big          & Standalone printer,  big coffee machine,  IKEA furniture (stool, utility cart, tray Table, nightstand) \\
\cmidrule{2-2}
Rare        & NavVis laser scanner,  ATV motor cycle,  wheel belt,  circuit breaker   \\        
\bottomrule
\end{tabular}}
\caption{\data includes 16 objects with diverse scales. Apart from common objects used in daily life, \data includes rare equipment from mechanical labs. 20 tasks are object-centric manipulation tasks for each object and the 4 IKEA furniture has both assembly and disassembly tasks.} \label{tab:objects}\vspace{-1em}
\end{table}
\vspace{-1mm}
\subsection{Data Collection and Statistics}
\vspace{-1mm}
\label{sec:data_collection_scenario}

\minisection{Tasks and objects.} We 
consider multi-step goal-oriented tasks 
involving 16 objects ranging from familiar 
objects often used in daily life to rare objects sometimes used in labs and 
factories as summarized in 
Table~\ref{tab:objects}. 
We consider small electronics like 
a {GoPro}, {DSLR camera}, and {Nintendo}, 
office
appliances like a {Nespresso machine} and 
{printer}, {IKEA furniture}, and objects in
labs such as a {laser scanner}, 
{motor cycle}, and {circuit breaker}. We have 
designed 20 tasks involving physical 
manipulation of these objects, \eg, 
changing batteries,  changing belts, 
furniture assembly, machine setup, etc. 
There is one task per object except for 
the IKEA furniture, which has assembly and 
disassembly tasks. Detailed task 
instructions are in supplementary 
materials.

\begin{figure}[t]
    \centering
    \includegraphics[width=\linewidth]{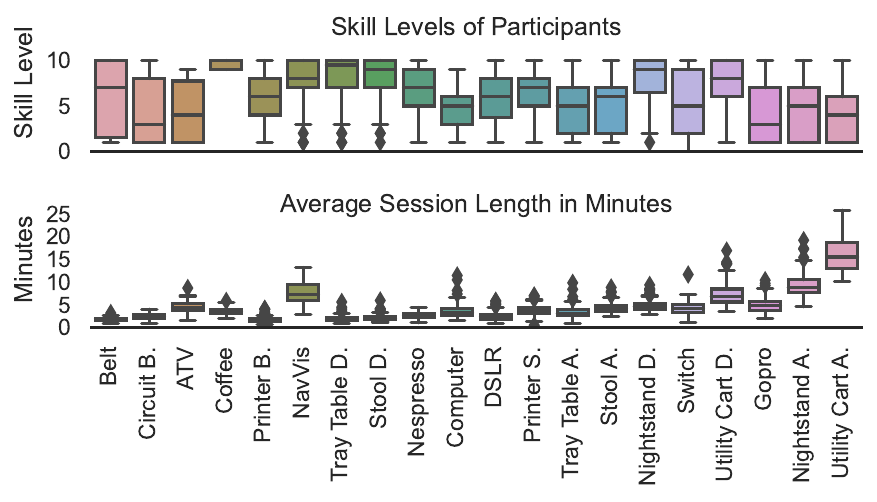}\vspace{-2mm}
    \caption{The skill level of participants (0-10) for the tasks is self-reported by the participants. The skill levels roughly reflect the length of the sessions though they might be noisy.}
    \label{fig:skill_level}
\end{figure}

\minisection{Participants and collection procedure.} We recruited 222 participants to form 350 unique pairs of instructors and performers for data collection. Figure~\ref{fig:demo_graphics} shows the demographics of the participants. Before data 
collection, the participants review the IRB forms to acknowledge the privacy and ethics standards (more 
details in supplementary 
materials). The instructors
are informed about the task in detail. The participants playing the role of
performers are only 
given a rough description of the tasks and scenarios beforehand and interacted with the objects based 
on their understanding. The instructors provide verbal guidance as the performers set out to complete the tasks. 

In Figure~\ref{fig:skill_level} (top), we show the distribution of the performers' familiarity with the tasks measured by a self-reported score (0-10) by the participants. We show the average length and the outliers of the recorded sessions in Figure~\ref{fig:skill_level} (bottom) to give a rough idea of how the participant's skill levels may lead to increased session variance. The participants' diverse skill levels and backgrounds provide rich information about the user behaviors and diverse interaction between the instructors and performers.

\minisection{Data capture tool.} We leveraged the Platform for Situated Intelligence framework \cite{bohus2021platform} to develop and open-source a distributed 
application for data capture using HoloLens 2~\cite{captureapp}. A client process running on the device captured the sensor data
while displaying a rectangular hologram 
frame around the user's visual field of 
view to guide their attention 
downward and keep the task actions in view 
of the sensors. Sensor data was streamed 
live over the network to a server  application that ran on a PC and persisted 
the data to disk. This distributed setup 
allows for collecting longer uninterrupted sessions without reaching the device storage 
capacity limits. 

\minisection{Comparison with other datasets.}  While there is no direct comparison of datasets with the same setup with \data, we list out different aspects of our dataset and compare it with related datasets in Table~\ref{tab:dataset comparison}.
\data is among the largest egocentric 
video datasets and features a multi-person 
collaboration setting, which is a unique 
addition to the field.  
In addition, \data relates to work on multi-agent collaborative simulation environments with a distinct characteristic of real-world sensor data and real-world human interaction. 

\begin{figure}[t]
    \centering
    \includegraphics[width=\linewidth]{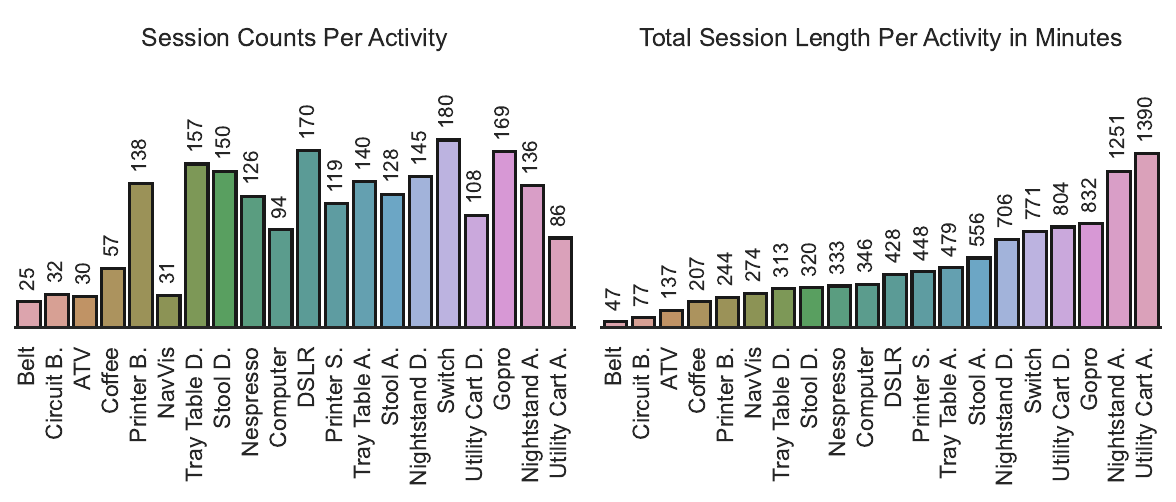}\vspace{-1mm}
    \caption{Data distribution of 166 hours captured by \data. \textbf{(left)} number of sessions per activity, and \textbf{(right)} total length of sessions in minutes.} 
    \label{fig:data_stats}
\end{figure} 

\begin{table}[t]
\centering
\adjustbox{width=\linewidth}{
\setlength{\tabcolsep}{2pt}
\def\arraystretch{1.2}
\begin{tabular}{@{}l|c|ccccccccc|cc@{}}
\toprule
Dataset          &  Settings
& \begin{tabular}[c]{@{}c@{}} Collaborative \\ $\&$Interactive \end{tabular}  & \begin{tabular}[c]{@{}c@{}} Instructional \\ $\&$Procedural\end{tabular}   & \begin{tabular}[c]{@{}c@{}} \# real \\ video hours\end{tabular}   \\ \midrule
Epic-Kitchen-100~\cite{epickitchen}           
& Cooking
&     \xmark             &  \xmark  & 100   \\
Assembly101~\cite{assembly101}    
& Toy assembly
&    \xmark       &  \cmark       &167     \\
Ego4D~\cite{ego4d}             
& Daily-life task
&  +     & +    & 3,670       
     \\
\midrule
VirtualHome~\cite{puig2018virtualhome}       
&  Household task

&      \xmark               &  \cmark &   \S     \\
ALFRED~\cite{ALFRED20}         
&  Household task &  \xmark   

     &       \cmark       & \S   \\
Habitat~\cite{szot2021habitat}      
&  Home assistance &\xmark   
&      \cmark           & \S      \\

BEHAVIOR~\cite{srivastava2022behavior}             
&     Daily-life task

 &  \cmark  &   \cmark  & \S    \\

IGLU~\cite{kiseleva2022interactive}      
&  \begin{tabular}[c]{@{}c@{}} Collaborative \\ building* \end{tabular} &\cmark   
    &  \cmark        & \S  \\
TEACh~\cite{padmakumar2022teach}           
&  Household task   &  \cmark&  \cmark

         & \S  \\

\midrule
\data (ours)    
&  Assistive task  
& \cmark            &\cmark     & 166     \\ 
\bottomrule
\end{tabular}}
\caption{\textbf{Comparison to related datasets and simulation platforms.} \data features a multi-person collaborative setting which is a unique addition to existing egocentric datasets in the real world. \data provides a set of instructional and procedure videos with multi-turn dialogues. Procedure tasks are defined as following a set of defined steps or procedures to achieve a specific goal, deviation from the procedure can be construed to be a mistake.
\data spans 166 hours and 2,221 sessions.  \S: simulation, *: Minecraft-like, +: partially included.
}
\label{tab:dataset comparison}
\end{table}

\vspace{-1mm}
\subsection{Annotations}
\label{sec:annotation}
\vspace{-1mm}

\begin{figure*}[t]
    \centering
    \includegraphics[width=\linewidth]{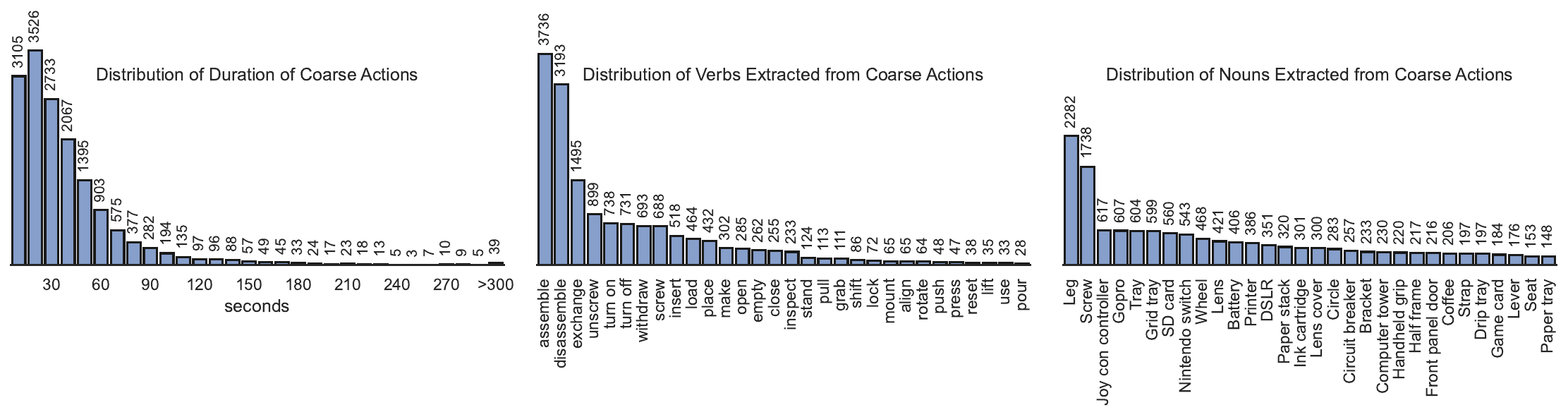}
    \caption{Data distribution of the coarse-grained actions. \textbf{(left)} duration of the actions in seconds, \textbf{(middle)} 30 most frequently occurring verbs, and \textbf{(right)} 30 most frequently occurring nouns.} 
    \label{fig:data_stats_coarse}
\end{figure*}

\begin{figure*}[t]
    \centering
    \includegraphics[width=\linewidth]{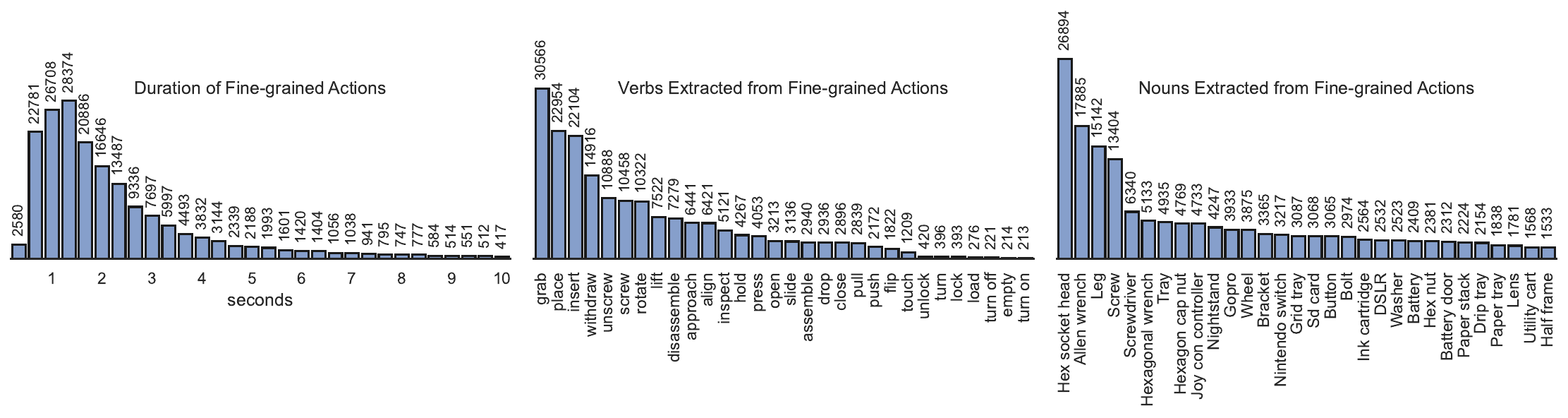}\vspace{-4mm}
    \caption{Data distribution of the fine-grained actions. \textbf{(left)} duration of the actions in seconds, \textbf{(middle)} 30 most frequently occurring verbs, and \textbf{(right)} 30 most frequently occurring nouns. We can see that most fine-grained actions last less than 2 seconds, and there is a long tail distribution in actions, verbs, and nouns.}
    \label{fig:data_stats_fine_grained} 
\end{figure*}

To better understand the actions and interactions in the dataset, we provide 
several sets of third-person manual annotations for text summaries, action 
segments, mistake attributes, and intervention attributes. 

\minisection{Language annotations.} We asked the annotators to watch the video and write a paragraph to describe the activities in the videos.  The description focuses on describing the hand actions in the procedure.
The third-person post hoc summary provides insights into the key moments during the interactive task completion. These could be used to build a comprehensive set of instructions for task completion.
We also provide the transcriptions of the conversations in the video.
With this set of annotations, we can understand the difference between third-person post hoc summaries and real-time conversations during the activities. More examples are in supplementary materials.  

\minisection{Coarse-grained action annotations.} The coarse actions usually 
describe a high-level step in the task (\eg, \emph{change battery of a GoPro} in the GoPro set up task) and can be divided into multiple fine-grained actions. To deal with the open-world setting, we ask the annotators to write a sequence  (\eg, \emph{man changes the battery of GoPro})  to describe the coarse actions and also identify the active verb-noun pair and optionally an adjective for the noun for benchmarking purposes (\eg, \emph{change battery}). 
The dataset 
includes 414 coarse-grained actions with 90 nouns and 39 verbs.
The distribution of the actions follows a long-tail distribution shown in Figure~\ref{fig:data_stats_coarse}, 
where 185 actions are considered head classes while the rest are considered tail classes according to the action frequency for evaluation purposes. 

\minisection{Fine-grained action annotations.} Fine-grained actions are the low-level atomic actions (\eg, \emph{press button}, \emph{grab screw}, etc.) for completing a step in the task, usually lasting for 1-2 seconds.
The fine-grained actions are presented in a
{verb-(adj.)-noun} pair format. There are 
1887 fine-grained actions with 165 nouns and 49 verbs.  
For a more comprehensive evaluation, we create a split of head 
actions with 1082 top actions and 805 tail actions. Distributions of fine-grained actions are shown in Figure~\ref{fig:data_stats_fine_grained}.  

As mentioned earlier, noun and verb vocabularies are not pre-defined but gradually built through annotation.  We ask the annotators to enter a new verb and noun if they cannot find it in the vocabulary. After the data is annotated, we ask the annotators to revisit and check the tail classes and see if they are repetitive to head classes.  Due to the open-world nature, some verb and noun combinations might be interchangeable with others in the list.  We show some examples in the supplementary materials.

\minisection{Mistake annotation.} Each fine-grained action is labeled as either \emph{correct} or \emph{mistake}, as indicated in Figure~\ref{fig:mistake_detection_example}. Mistakes include the ones that are ``self-corrected by the 
task performers'', are ``verbally corrected by 
the instructors'', and ``are not corrected labeled''. Our human annotators annotate all three mistake types separately, but for benchmark evaluation, we will consolidate them into one mistake class. We defer the detailed study of differentiating whether and how the mistakes are corrected to future work.  To ensure the annotation quality, 
we additionally ask the third-person annotators to explain why the action is a mistake and also assign a mapping to every mistake that is corrected by an instructor verbally to the conversation sentence whose type is ``instructor correcting mistakes''.

\minisection{Intervention 
annotation.}  Since instructors assist the 
task performers verbally, we 
annotate the conversation between the 
instructors and performers to reflect the 
interventions in task completion. We annotate each conversation sentence with two attributes to indicate the conversation types and the conversation initiator. The conversation initiators can either be the ``task performer'' or the ``instructor''. And the sentence purpose types can be the 
instructor ``correcting 
mistakes'', ``answering questions'', 
``following up with more instructions'', ``confirming previous actions''. ``describing the high-level task'', ``opening/closing remarks'', or the task performer starting the conversation to ask questions.  
The human annotators watch the videos and use their best judgment to 
annotate the roles of different conversations in the videos.  

In our benchmarks, we consider 3 
intervention types: \emph{correcting mistakes}, \emph{following up with more
instructions}, and 
\emph{confirming the previous action} as they are 
more related to physical actions in the task procedure. Figure~\ref{fig:mistake_detection_example} shows examples of conversational interventions.  More examples can be found in the supplementary materials. 

\minisection{Audit process.}
Annotations are done by professional annotators based on the following process. The annotators first take a pass on the video to add the fine-grained actions, coarse-grained actions, conversation, and text summary, along with the associated annotation elements for each event. After self-review, the annotated data is passed to an independent reviewer for auditing, and the mistakes are fixed directly or sent back to the original annotator for updates. The annotations finally go through a targeted review to check the open-ended text fields like narration, action sentences, and conversation transcriptions to ensure consistency. Before the annotations were delivered,  we applied a list of constraints to systematically check the annotations to further dig out the wrong annotations. 

\begin{table}[t]
\centering
\adjustbox{width=.9\linewidth}{
\begin{tabular}{@{}cc@{}}
\textsc{Immediate intervention} & \textsc{Lazy intervention}   \\ \midrule
\begin{tabular}[c]{@{}l@{}}insert screw\\ approach button\\ insert joy con controller\\ place tray\\ pull battery door\end{tabular} & \begin{tabular}[c]{@{}l@{}}drop allen wrench\\ drop hex socket head\\ drop screw\\ screw hex socket head\\ screw screw\end{tabular}  \\
\bottomrule
\end{tabular}}
\caption{Top mistakes that are corrected immediately \textbf{(left)} or later and sometimes through self-correction \textbf{(right)}.}. \label{tab:mistake_intervention_correlation} \vspace{-4mm}
\end{table}
\vspace{-3mm}
\section{Observations and Tasks from \data}
\vspace{-1mm}
Here we present observations from \data in 
Section~\ref{sec:observations} and identify a few open problems 
that are necessary components for an interactive AI 
assistant. In Section~\ref{sec:benchmark_design}, we define new 
tasks and benchmarks with \data. 

\subsection{Observations from \data}
\label{sec:observations}

\minisection{Correlation between mistakes and intervention.} We find that the response time for the 
human assistant to intervene in the procedure and correct mistakes depends on the severity level of the mistakes. If
mistakes are critical, the instructors proactively interrupt the student immediately (within less than 5
seconds) while other mistakes are either self-corrected by the task performer or corrected later by the 
instructor.  

 In Table~\ref{tab:mistake_intervention_correlation}, we present the top actions that need immediate intervention and the 
 top actions that are self-corrected by the task performers or 
 corrected later by the instructors.  We notice that actions 
 related to linear tasks, where the task progression is stalled 
 if steps are not followed in order, are often intervened 
 immediately by the instructors. For example, ``insert joy con controller", and ``place tray" etc. In 
 contrast, the lazily edited corrections are related to furniture assembly such as the actions ``drop allen wrench", 
 ``drop hex socket head", etc. These are tasks where mistakes are 
 unclear in every stage, and the user often 
 intuitively adjusts their steps.

\begin{table}[t]
\centering
\begin{tabularx}{\columnwidth}{>{\raggedright\arraybackslash}X}
\textbf{\small{Spatial deixis from the intervention transcripts}} \\
\midrule
\small{``You should press the button that's on the body of the camera just \underline{at the right of the lens."}}\\
\addlinespace
\small{``You should leave the bolt, \underline{like it was before}."}\\
\addlinespace
\small{``The button is \underline{on the other side} of the Switch."}\\
\addlinespace
\small{``The SD card comes \underline{from the right slot}, \underline{on the right hand side} and it opens by using a knob \underline{next to the screen} \underline{on the right bottom side} of the screen."}\\
\addlinespace
\small{``Currently the tray is \underline{facing you}, please rotate so the back is facing you."}\\
\addlinespace
\small{``You should put it \underline{the other way around}. It's upside down."}\\
\addlinespace
\small{``Please start by removing the screw \underline{of the top shelf} first."}\\
\addlinespace
\small{``And now, ehm, it is, it should be \underline{the other one} that should be \underline{on top of the other}."}\\
\addlinespace
\small{``\underline{To the right of that}, there is a tiny little square."}\\
\bottomrule
\end{tabularx}
\caption{Examples of deictic phrases that help in grounded guidance by specifying contextual spatial locations.} 
\label{tab:spatial-deictics}
\end{table}

\minisection{Grounded guidance.}
We also notice that the instructions from human instructors are often grounded in the 3D environment. 
An important aspect of grounded guidance is the ability to communicate about the physical world by pointing to things in
context. Spatial deixis refers to phrases 
that are used to locate things in space and to express direction 
and distance. The deictic analysis of the transcripts in \data 
reveals a wide set of words that indicate specific 
and relative locations and directions, especially during 
interventions to correct mistakes. A list of some prominent 
examples is shown in Table~\ref{tab:spatial-deictics}.  

\vspace{-2mm}
\subsection{Benchmark Tasks}
\vspace{-1mm}
\label{sec:benchmark_design}

Inspired by the observations above,  we think it is important for an 
interactive AI assistant to have a good world state estimation model that 
can detect mistakes and predict whether to intervene in the task 
procedure. Besides, augmenting instructions with spatial guidance can  be useful for AI agents. To this end, we introduce new mistake detection, 
intervention prediction and  3D hand pose forecasting tasks for 
interactive and grounded guidance. 
Additionally, we benchmark models on action recognition tasks following the convention in ~\cite{epickitchen}. 

\minisection{Mistake detection} is defined following the convention~\cite{assembly101} but applied to fine-grained actions in our benchmark. We take the features from the fine-grained action clips from the beginning of the coarse-grained action until the end of the current action clip, and the model predicts a label from \emph{\{correct, mistake\}}. The task is challenging given that the class distribution is highly skewed, with around 6\% mistakes among the fine-grained actions. 

\minisection{Intervention type prediction} is to predict the intervention types given an input of a window of 1, 3, or 5 seconds before the intervention. This newly proposed benchmark is to test if the model can correctly figure out the correct intervention types during task completion. 
Currently, \data includes 3 intervention types, and we report the precision and recall of each intervention type. 

\minisection{3D hand pose forecasting} is another new benchmark 
introduced by \data. Existing action forecasting work~\cite{epickitchen} 
mostly focuses on providing semantic labels of 
future actions and does not provide explicit 3D guidance on hand poses. 
Predicting 3D hand poses can be useful for various 
applications~\cite{saycan}, and it can augment instructions and 
spatially guide users in different tasks. 
In this benchmark, we take 
3 seconds inputs similar to other 3D body location forecasting literature~\cite{zheng2022gimo} and forecast the continuous 3D hand poses for the next 0.5, 1.0, and 1.5 seconds. The evaluation metric is 
the average of mean per joint position error over time in centimeters compared to ground truth. 
To have a proper evaluation metric that can help 3D action guidance, we remove the mistakes from the action sequences and only forecast 3D hand pose for the correct labels.

\section{Experiments}
\label{sec:benchmark}

\begin{table*}[t]
\centering
\setlength{\tabcolsep}{4pt} 
\adjustbox{width=\linewidth}{
\def\arraystretch{1.2}
\begin{tabular}{@{}l|c|ccc|ccc|ccc@{}}
\toprule
&\multirow{2}{*}{Mods.} & \multicolumn{3}{c|}{All Classes Accuracy}& \multicolumn{3}{c|}{Head Classes Accuracy} & \multicolumn{3}{c}{Tail Classes Accuracy}               \\
 & & Top1 / 5 Act  & Top1 / 5 Verb  & Top1 / 5 Noun  & Top1 / 5 Act  & Top1 / 5 Verb  & Top1 / 5 Noun  & Top1 / 5 Act  & Top1 / 5 Verb  & Top1 / 5 Noun  \\\midrule
\multirow{5}{*}{\rotatebox{90}{Pretrained}}&RGB &  34.83/68.60 &	42.14/78.96	& 66.81/90.04 &	35.26/69.34 &	42.56/79.53 &	67.19/90.36 &	0.03/0.17	& 10.86/36.33 &	38.01/66.48 \\
&Hands & 20.86/43.92 & 35.38/65.76 & 37.10/63.42 &	21.13/44.50	& 35.72/66.30 & 37.50/64.06 & 0.00/0.01	& 10.11/25.47 &	7.30/15.54 \\
&R+H & 35.06/68.95 & 42.45/79.42 &	67.05/90.01	& 35.49/69.71 &	42.87/80.03 & 67.43/90.32 &	0.03/0.16 &	11.05/33.71	& 38.39/66.85 \\
&R+H+E & 35.27/68.69 & 42.92/79.11 & 67.03/89.96 & 35.72/69.42 & 43.33/79.67 & 67.45/90.29 &	0.03/0.18 &	11.99/37.64	& 35.96/65.17 \\
&R+H+E+I & 34.80/68.26 & 42.24/78.88 & 66.65/89.76 &35.23/69.00 &	42.63/79.46 & 67.03/90.08 &	0.03/0.17 & 12.92/35.96	& 38.20/65.17 \\
\midrule 
\multirow{3}{*}{\rotatebox{90}{Scratch}} & RGB & 18.78/48.09 & 28.45/65.43 & 43.72/73.69 & 19.03/48.72	& 28.74/66.00 &	44.09/74.21 &	0.00/0.01	& 6.55/22.66 & 15.73/34.27 \\
& Hands & 23.94/47.58 &	39.79/68.76	& 39.34/65.76 &	24.26/48.21	& 40.18/69.30 & 39.79/66.43 &	0.00/0.00 &	10.67/28.65 &	5.62/16.29  \\
& R+E & 20.86/50.27 &	30.92/67.59 &	45.28/75.29	& 21.13/50.93 &	31.22/68.20 &	45.65/75.80	& 0.00/0.01 &	8.05/21.72 &	17.60/37.08 \\
& R+H & 29.32/59.20	& 41.48/73.92 &	52.54/80.65 &	29.70/59.95 &	41.87/74.51 &	52.99/81.17 &	0.00/0.03 &	12.55/29.78	& 19.29/41.76 \\
& R+H+E & 29.58/59.14 &	41.58/73.73	& 52.65/80.76 &	29.97/59.89	& 41.99/74.31 &	53.10/81.29 & 0.00/0.04	& 10.67/29.96	& 19.29/40.82 \\
& R+H+E+I & 26.87/56.13 &	39.29/72.49	& 49.96/78.60	& 27.22/56.82 &	39.69/73.06 &	50.35/79.07 &	0.01/0.05 &	8.99/29.96	& 21.16/43.26 \\
\bottomrule
\end{tabular}}
\caption{\textbf{Fine-grained action recognition benchmark results.} We report baseline results of (multimodal) TimeSformer~\cite{timesformer} models on \data. Here \texttt{R}, \texttt{H}, \texttt{I} denote RGB, hands, and head pose (whose sensor source is similar as IMUs). When starting from a pre-trained ViT model on ImageNet, all the models have higher performance than training from scratch. However, with a pre-trained image model, the influence of other 
modalities is reduced. If the models are trained from scratch, we can see that adding hands can improve the prediction of \texttt{verbs} and lead to better results than the baseline \texttt{RGB} only model. Simply concatenating more sensors as inputs may not necessarily lead to improved performance. } \label{tab:action_recog} \vspace{-2mm}
\end{table*}
\begin{table*}[t]
\centering
\setlength{\tabcolsep}{4pt} 
\adjustbox{width=\linewidth}{
\def\arraystretch{1.2}
\begin{tabular}{@{}l|c|ccc|ccc|ccc@{}}
\toprule
&\multirow{2}{*}{Mods.} & \multicolumn{3}{c|}{All Classes Accuracy}& \multicolumn{3}{c|}{Head Classes Accuracy} & \multicolumn{3}{c}{Tail Classes Accuracy}               \\
 & & Top1 / 5 Act  & Top1 / 5 Verb  & Top1 / 5 Noun  & Top1 / 5 Act  & Top1 / 5 Verb  & Top1 / 5 Noun  & Top1 / 5 Act  & Top1 / 5 Verb  & Top1 / 5 Noun  \\\midrule
\multirow{5}{*}{\rotatebox{90}{Pretrained}}&RGB &  50.91/86.89	& 60.51/93.45 &	73.35/95.90	& 53.40/89.53	& 62.78/95.39 &	75.00/96.70 &	0.37/2.24 & 20.00/58.89 &	43.89/81.67\\
&Hands & 22.20/54.16 &	35.12/72.34	& 37.43/68.33 &	23.44/57.08	& 36.75/75.25 &	38.84/70.45 &	0.00/0.12	& 6.11/20.56	& 12.22/30.56 \\
&R+H &  50.80/86.54 & 59.71/93.36 &	73.20/95.78	& 53.27/89.12 &	61.94/95.20 &	75.09/96.57	&0.37/2.28 &	20.00/60.56	& 39.44/81.67 \\
&R+H+E & 50.35/85.51 &	58.85/93.21	& 73.67/95.63 &	52.65/88.03 &	60.85/94.89 & 75.28/96.32 &	0.53/2.28 &	23.33/63.33 &	45.00/83.33 \\
&R+H+E+I & 50.18/86.07 &59.00/93.21	& 73.64/96.25 &	52.49/88.78 &	61.22/95.07 & 75.41/96.91 &	0.50/2.12 &	19.44/60.00	& 42.22/84.44 \\
\midrule 
\multirow{3}{*}{\rotatebox{90}{Scratch}} & RGB & 32.17/70.45 &	41.88/84.77 &	53.81/84.12 &	33.92/73.75 &	43.49/87.34	& 55.42/85.97 &	0.06/0.65 &	13.33/38.89 &	25.00/51.11\\
& Hands & 27.36/59.71 &	40.17/76.33	& 43.77/73.64 &	28.83/62.78	& 41.74/78.90 &	45.36/75.75 &	0.06/0.28 &	12.22/30.56	& 15.56/36.11 \\
& R+E & 31.32/69.86 &	41.71/84.36	& 53.81/83.68 &	32.92/73.04	& 43.20/86.94 &	55.42/85.41 &	0.16/0.75 &	15.00/38.33	& 25.00/52.78 \\
& R+H & 34.42/73.02 &	45.28/85.09 &	56.14/86.45	& 36.16/76.15 &	47.04/87.44	&57.76/88.28 &	0.19/0.97 &	13.89/43.33 &	27.22/53.89 \\
& R+H+E & 35.18/73.91 &	45.51/85.80 &	56.85/86.57 &	37.13/77.12 &	47.54/88.37 &	58.70/88.40 &	0.03/0.94 &	9.44/40.00	& 23.89/53.89\\
\bottomrule
\end{tabular}}
\caption{\textbf{Coarse-grained action recognition results.} The overall trend is similar to fine-grained action recognition.} \label{tab:coarse_action_recog} \vspace{-2mm}
\end{table*}
In this section, we provide the evaluation results
of the proposed benchmarks. We will start with
the standard action recognition benchmarks, and then we will 
present the results of the newly 
proposed benchmarks.  We also provide 
ablations of different sensor modalities to understand the roles of different sensors in various tasks. 
We hope the baseline results can guide future research in this space. 

\minisection{Implementation details.} We adopt 
TimeSformer~\cite{timesformer}, a state-of-the-art  vision transformer (ViT)~\cite{vit} based video model, as the backbone and change the head with a different number of classes for different benchmarks. 
We modify the original TimeSformers to perform multimodal learning by introducing additional tokens for different modalities and embedding layers to encode the additional sensor modalities.  Specifically, we use 26$\times$2 tokens for both left and right hands (one token for one hand joint),  one token for eye gaze, one token for head poses, and  196 tokens for depth.  We can enable and/or disable different modalities during training and evaluation.  Detailed configurations are available in supplementary materials.  Note that we consider the resulting model a vanilla multimodal model and serve as a baseline for future studies.  

We randomly split the 2221 sessions into train, validation, and test sets following a ratio of 70\%, 10\%, and 20\% on a per-task basis, which includes 1545 sessions for training, 213 sessions for validation, and 463 sessions for testing.  We also synchronize all the modalities according to the video stream and keep the frame rate at 30 fps by sub-sampling other modalities in our experiments. 

During training, the model is trained for 15 epochs with an initial learning rate of 0.01 and a batch size of 64 using stochastic gradient descent. We divide the learning rate by 10 at the epoch 11 and 14. For each input segment, we randomly sample 8 frames/data points within the segment. We train our models with 4xA6000 GPU machines, and the fine-grained action recognition runs usually take about one day. We also train the model with random initialization and an ImageNet pre-trained ViT backbone.  

For our hand pose forecasting benchmark, which is a regression task, we adopt Seq2Seq model~\cite{sutskever2014sequence} following~\cite{liu2022joint}. 

\begin{figure*}[t]
    \centering
    \includegraphics[width=\linewidth]{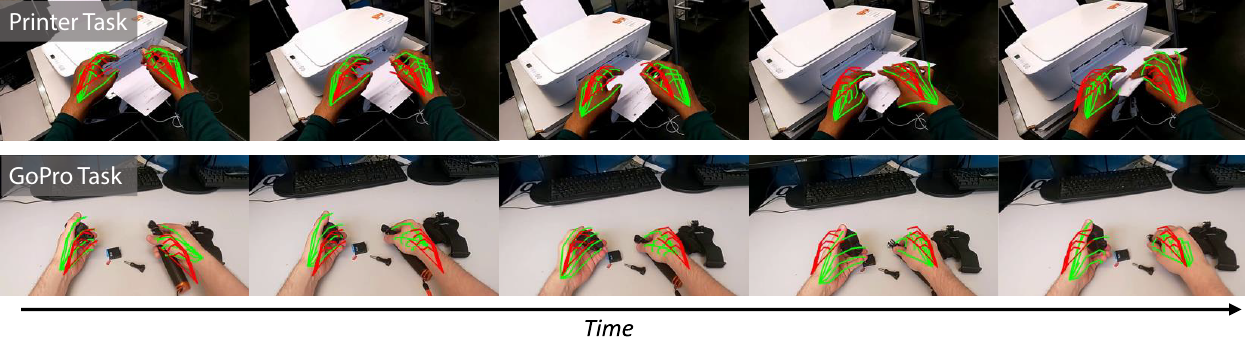} \vspace{-4mm}
    \caption{\textbf{Qualitative visualization for 3D hand pose forecasting}. We visualize the ground-truth hand joint positions (\ie, hand pose, visualized in \textcolor{green}{Green}) and the prediction of the hand pose for the next 1.5 seconds (visualized in \textcolor{red}{Red}). The input to the model is a 3-second long clip ahead of the prediction. The task is challenging as hands often move quickly. As we can see from the figure,  the predicted hand pose is more off from the grounded truth in the longer future.}
    \label{fig:3d-hand-guidance-qualitative} \vspace{-2mm}
\end{figure*}

\begin{table}[t]
\centering
\def\arraystretch{1.1}
\adjustbox{width=.9\linewidth}{
\begin{tabular}{@{}lccccc@{}}
\toprule
\multirow{2}{*}{Mods.} & \multirow{2}{*}{F-score}  & \multicolumn{2}{c}{Correct} & \multicolumn{2}{c}{Mistake} \\
& & Precision & Recall & Precision & Recall \\
\midrule
Random & 27.71  & 60.87 & 10.22 & 15.00 & 46.15 \\\midrule
RGB & 35.11 & 82.56 & 51.82 & {12.96} & 26.92  \\
Hands & 40.19  & {92.68} & {52.41} & {12.5} & {31.25} \\
\midrule
R+H & 36.18 & 85.51 & 43.07 & 9.68 & 11.54 \\
R+H+E & 32.08 & 88.57 & 42.76 & 11.43 & 50.00\\
\bottomrule
\end{tabular}
}
\caption{\textbf{Mistake detection results.} Similar to action recognition, adding hand poses to RGB (\texttt{R+H}) improves an RGB-only model from 35.11 to 36.18 points. What's different is that we find the \texttt{hands} only model outperforms other modalities in mistake detection, achieving 40.19 points.} \label{tab:mistake_detection_table} \vspace{-4mm}
\end{table}

\minisection{Action recognition.} We show the fine-grained action recognition results in Table~\ref{tab:action_recog} and the coarse-grained action recognition results in Table~\ref{tab:coarse_action_recog}. We can see that initialized with a pre-trained ViT model on ImageNet; all models can achieve around 35\% top-1 accuracy on fine-grained action recognition and 50\% top-1 accuracy on coarse-grained action recognition. This is comparable to baselines of other 
fine-grained action benchmarks~\cite{assembly101} (23\% top-1 accuracy),

We notice that the pre-trained model on ImageNet may reduce the influence of other modalities that are not pre-trained. If trained from scratch, we can see from Table~\ref{tab:action_recog} (bottom) that adding hands can improve the prediction of verbs and lead to better results than the RGB-only model. 

\minisection{Mistake detection.} In Table~\ref{tab:mistake_detection_table}\footnote{\label{note1}Evaluated on 10\% of the entire data }, we present the results of mistake detection where the features are extracted from the pre-trained action recognition models shown in Table~\ref{tab:action_recog}. Here we find the hand poses information benefits the task and outperforms the other modalities.

\begin{table}[t]
\centering
\setlength{\tabcolsep}{2pt} 
\adjustbox{width=\linewidth}{
\def\arraystretch{1.2}
\begin{tabular}{m{0.16\linewidth}|cc|cc|cc|cc}
\toprule
 \multirow{2}{*}{Mods.} & \multicolumn{2}{c|}{Overall} & \multicolumn{2}{c|}{Confirm Action} & \multicolumn{2}{c|}{Correct Mistake} & \multicolumn{2}{c}{Follow-up} \\
 &Prec. & Recall & Prec. & Recall & Prec. & Recall & Prec. & Recall \\ \midrule
 RGB & 47.92 &	46.09 &	44.09 &	27.65 &	46.13 &	41.44 &	53.54 &	69.18 \\
 Hands & 15.75 &	33.33 &	0 &	0 &	0 &	0 &	47.25 &	100 \\
 R+H & 48.08 &	47.06 &	43.64 &	32.21 &	46.63 &	44.67 &	53.95 &	64.3 \\
 R+E & 48.33	& 47.38 &	45.45 &	31.87 &	45.16 &	45.16 &	54.39 &	65.1 \\
 R+H+E+I & 48.75	& 47.75 &	45.55 &	33.94 &	46.11 &	44.91 &	54.59 &	64.42 \\
\bottomrule
\end{tabular}} \caption{\textbf{Intervention type prediction results.} The classes in the benchmark are highly skewed. We can see that adding hands and eyes improves the intervention-type prediction.\label{tab:intervention}}
\end{table}

\minisection{Intervention type prediction.} For intervention prediction, we show the results in Table~\ref{tab:intervention}\footref{note1}. We can see adding hands and eye gaze (\texttt{R+H+E}) can significantly boost the overall precision and recall to 48.31\% and 37.59\%, improving about 35 and 4 percentage points over \texttt{RGB}. This may be because eye gaze is a forecasting signal, as people often look at the regions before the action starts, which can assist the models to attend to important regions for better anticipation.

\begin{table}[t]
\centering
    \setlength{\tabcolsep}{12pt} 
    \def\arraystretch{1}%
    \adjustbox{width=.9\linewidth}{
    \begin{tabular}{m{0.3\linewidth}ccc}
    \toprule
    \multirow{2}{*}{Mods.} & \multicolumn{3}{c}{Mean Error Distance (cm, $\downarrow$)}\\ 
     & 0.5 sec & 1.0 sec & 1.5 sec\\ \midrule 
    Static-H & 9.34 & 13.91 & 16.70 \\ \midrule
    Hands &  9.80 & 10.68 & 11.25 \\ 
    \midrule
    H+E &  9.80 & 10.70 & 11.25  \\
    H+E+I & 9.80 & 10.69 & 11.25  \\
    R+H+E & 9.73  & 10.65  & 11.22  \\
    R+H+E+I & 9.72 & 10.62 & 11.19 \\
    \bottomrule
    \end{tabular}}\caption{\textbf{3D hand pose forecasting benchmark results.} We report the mean per joint position error for 0.5  1.0, 1.5 seconds (lower the better). The static hand baseline (\texttt{Static-H}) refers to using the last frame of the input. The Seq2Seq~\cite{sutskever2014sequence} model trained using hands only \texttt{H}, and using only hand \texttt{H} achieves better results than \texttt{Static-H}.}\label{tab:hand_pose} \vspace{-5mm}
\end{table}
\minisection{3D hand pose forecasting.}  Table~\ref{tab:hand_pose} shows that given only the 3D poses of hands as input, the model can perform with the accuracy of 9.80, 10.68, and 11.25 centimeters,  
the average of mean per joint position error, for 0.5, 1, and 1.5 seconds, respectively. We should note that this task is challenging as hands can move quickly within a window of 0.5 seconds. Compared to \texttt{static-H}, which uses the last 3D hand pose, our baseline (\texttt{H}) already outperforms it.
In Figure~\ref{fig:3d-hand-guidance-qualitative}, we show the visualization of the hand pose forecasting. 

\minisection{Importance of hand poses and eye gaze.} As we can already see from the Tables~\ref{tab:action_recog}, ~\ref{tab:mistake_detection_table}, ~\ref{tab:intervention}, ~\ref{tab:hand_pose}, the 3D hand poses and 
eye gaze can help the model prediction to recognize  actions, detect mistakes, and understand users' 
intentions. These modalities can augment the commonly used RGB images for better performance.  Simply concatenating more modalities (e.g., depth, head poses) as 
inputs may not necessarily lead to a more capable model, as those modalities may need specialized encoders or model architectures to process them simultaneously. 
We believe \data will enable and foster further research in multi-modal learning in this direction.

\section{Conclusion and Future Work}
\label{sec:conclusion}
In this work, we identified and explored several important problems with building an interactive AI assistant in the 
physical world.  We introduced a large-scale multimodal 
egocentric video dataset, \data, containing rich 
information about human interaction in an assistive task 
completion setting. The task performer wears a HoloLens 2 
headset while completing various object-centric 
manipulation tasks. The real-time video feed from the headset is sent to a remote instructor who provides verbal guidance to the task performer. \data 
captures seven raw sensor modalities during the interaction, and among them, we found hand pose and eye gaze are useful information sources for an interactive 
AI agent.  By augmenting the data with additional third-person manual annotations on action segments, mistakes, 
and intervention types, we constructed new benchmarks on 
mistake detection, intervention type prediction, and 3D 
hand pose forecasting, which we believe is a necessary 
component for an interactive and grounded AI assistant. 
As a first step in this direction, this work also leaves room for future work to improve upon (\eg, annotating object poses in the data, investigating object-centric models of affordance and manipulations in AI assistance, etc.). We believe \data, coupled with the associated benchmarks and tooling will benefit future research into building competent AI assistants for everyday tasks in the real world.

\subsection*{Acknowledgement}
\small{
We thank all the 222 data collectors who participated in the study and acknowledge the hard work of the annotation team led by Megan Yuan and Dan Luo from DataTang Technology Inc. We thank Nick Saw for the help in the data collection software and  Yale Song and Vibhav Vineet from the Computer Vision Group at Microsoft Research for the early discussion. We also thank the feedback from colleagues at MSR. }

{\small
\bibliographystyle{ieee_fullname}
\bibliography{reference}
}
\newpage

\newcommand{\beginsupplement}{%
        \setcounter{table}{0}
        \renewcommand{\thetable}{S\arabic{table}}%
        \setcounter{figure}{0}
        \renewcommand{\thefigure}{S\arabic{figure}}%
        \setcounter{section}{0}
        \renewcommand{\thesection}{S}%
     }

\onecolumn
\section*{Supplementary Material}
\beginsupplement

This supplementary material shows additional qualitative results and sample visualization of our dataset. Also, we show details about our annotation method, data capture procedure, and data analysis. Furthermore, we will report implementation details, and qualitative results. Additional qualitative results and visualization of our dataset can be found at \small{\url{https://holoassist.github.io/}}.

\subsection{Interactive Assistive Task Completion}
\begin{figure*}[h]
    \centering
    \includegraphics[width=.9\linewidth]{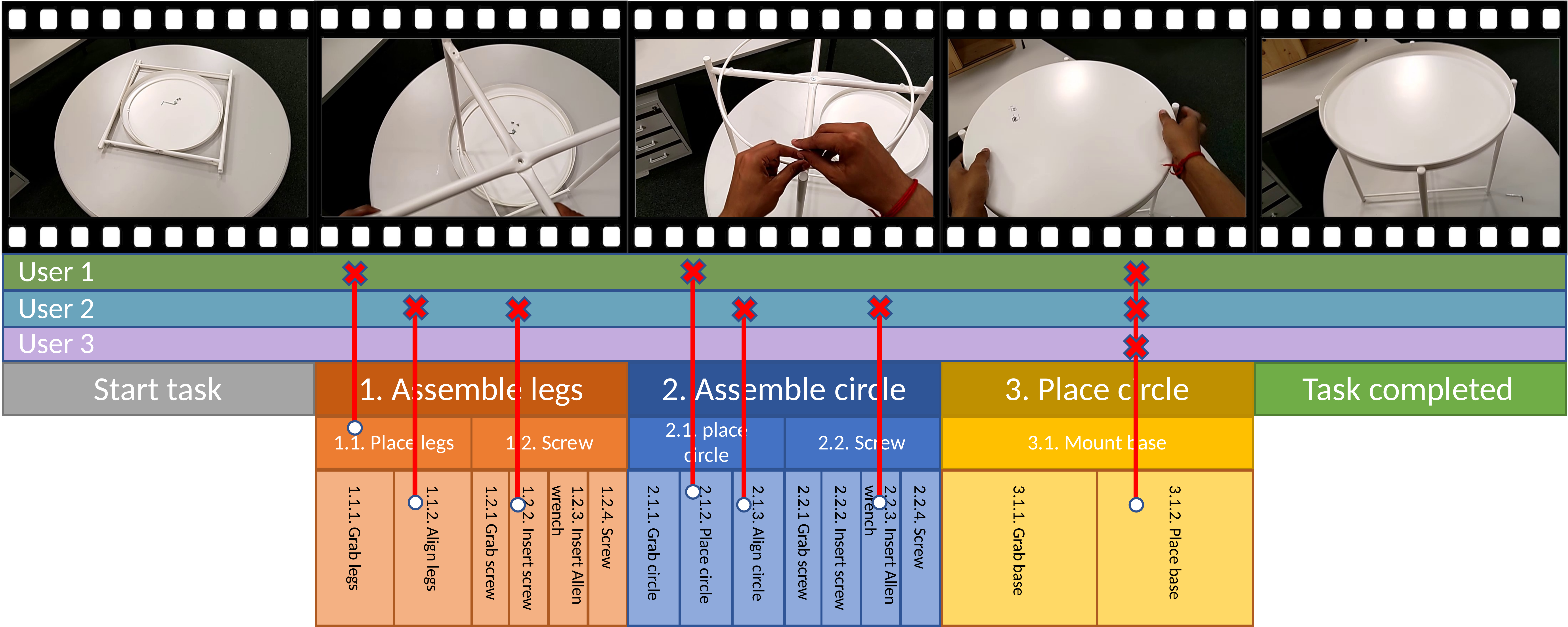}
    \caption{Intervention-actions structure. This figure shows the tree structure of different levels of structures (high, mid, or low-level instructions) for the \texttt{Assemble tray table} task, together with the intervention moments indicated by \texttt{red crosses} for three different \textit{task performers}. As the figure illustrates, different task performers need different levels of supervision at different moments, which shows the importance of our dataset for building an interactive AI assistant. }
    \label{fig:internvention-tree}
\end{figure*}

\data features a unique interactive assistive task completion setting where instructors intervene during the task completion process if task performers make mistakes or get confused. In Figure~\ref{fig:internvention-tree}, we show additional illustrations of how different users may experience 
different intervention moments while completing the same task. The diversity and complexity of the intervention moments and types suggest the challenges of building an interactive AI assistant.  

\subsection{Data}
\subsubsection{Instructor Training}
Before data collection starts, we go through a training process for the instructors to get familiar with the objects and tasks. We provide the instructors with detailed guidelines for the example tasks as shown in Table~\ref{tab:task_instructions}. The steps in the task instructions are mainly used as a reference and the instructors do not need to strictly follow the suggestions. They can change the order, add or skip tasks, and change the way they instruct the process as they see fit during the data capture. 

\begin{longtable}{m{0.15\linewidth}|p{0.5\linewidth}|p{0.3\linewidth}}
\toprule
Objects & Example Tasks & Comments \\
\midrule
GoPro & \begin{tabular}[c]{@{}p{\linewidth}@{}} - Change Battery\\ - Change micro sd card\\ - Turn on \\ - Turn off\\ - Put on a strap\\ - Change the strap to handheld grip (or tripod)\\ - Remove the handheld grip (or tripod)\end{tabular} &  \begin{tabular}[c]{@{}p{\linewidth}@{}}Turning on/off is controlled by the button on the side of the GoPro not the top of device and you need to press $\sim$3 seconds to make it happen.\end{tabular} \\ \midrule
Nintendo & \begin{tabular}[c]{@{}p{\linewidth}@{}}- Change controller from the handheld mode to pad/joy-con comfort grip\\ - Change controller from pad/joy-con comfort grip to two joy-cons \\ - Change controller from two joy-cons back to the handheld mode\\ - Stand Nintendo using kick stand\\ - Change game card\\ - Change micro sd card\\ - Turn on \\ - Turn off (hold for 3 seconds and power off) - just one second is only sleeping mode\end{tabular} & \begin{tabular}[c]{@{}p{\linewidth}@{}}
1. The task always starts from the handheld mode. Also, you need to turn off the device entirely before starting the data capture.\\ 
2. For two joy-cons, task performers need to lock them using the white button. \\
3. The strip always faces towards the wrist side.\end{tabular} \\ \midrule
DSLR & \begin{tabular}[c]{@{}p{\linewidth}@{}}- Detach lens\\ - Attach lens\\ - Detach a lens cover\\ - Attach a lens cover\\ - Turn on \\ - Turn off\\ - Change battery\\ - Change SD card\end{tabular} &
\begin{tabular}[c]{@{}p{\linewidth}@{}}
1. When you turn on the camera, please wait until the screen is on so we can capture the visual cues. \\
2. For some DSLR models, make sure to put a microSD card in the SD card adapter - or the camera may not be turned on properly. \end{tabular} \\ \midrule
Espresso Machine & \begin{tabular}[c]{@{}p{\linewidth}@{}}- Add coffee beans\\ - Make a cup of coffee\\ - Add milk/cream and stir\\ - Empty drip tray\end{tabular} &  \\ \midrule
Nespresso / Capsule Machine & \begin{tabular}[c]{@{}p{\linewidth}@{}}- Please turn on the machine before starting capture\\ - Add water\\ - Make a cup of coffee (put a capsule) - make sure the capsule went to the capsule container\\ - Empty the capsule container into the trashcan  \\ - Empty drip tray\end{tabular} & The coffee is hot so please make sure that the participants dispose the waste into the sink can directly. \\ \midrule
Printer (Big) & \begin{tabular}[c]{@{}p{\linewidth}@{}}- Wake up from sleep mode (Turn on)\\ - Go into sleep mode (Turn off)\\ - Add/load paper to tray 1 (Please say tray 1 explicitly)\\ - Change black printer cartridge\end{tabular} &  \\ \midrule
\begin{tabular}[c]{@{}l@{}}Printer\\ (small)\end{tabular} & \begin{tabular}[c]{@{}p{\linewidth}@{}} - Load paper and pull out the tray.\\ - Copy a black-and-white paper (make sure the subject removes the paper at this stage)\\ - Change (install) color and black printer cartridge  (Don’t turn off the printer at this stage)\\ - Close everything\end{tabular} & 
\begin{tabular}[c]{@{}p{\linewidth}@{}}
1. Reset the machine first.  \\
2. Details for steps 1 and 3 are on Page 4 of the manual. You can instruct the process according to the manual. \\
3. If you can hear noise after changing the cartridge, please ignore it and just close everything. \\
4. When you start, the power button should not be blinking. Press the cancel button if things go wrong.\\
5. When copying the paper, please place the paper on the right-bottom of the machine. \end{tabular} \\ \midrule
Small furniture from IKEA & \begin{tabular}[c]{@{}p{\linewidth}@{}}For IKEA furniture, please follow the steps in the manual.  \\ - Assemble\\ - Disassemble \end{tabular} &  Assemble and disassemble should be in different videos \\ \midrule
Objects in the lab & Instructors decide the setup procedures or steps for manipulating the objects. & \\
\bottomrule
\caption{Task instructions for instructors to get familiar with the objects. Instructors are free to add or skip tasks, change orders, and use their own way of completing the tasks during the data collection.} \label{tab:task_instructions}
\end{longtable}

\subsubsection{IRB Approval for Data Collection}
The data collection process of \data was reviewed and approved by IRB before the work started.  The data capture mostly takes place in public areas (\ie, offices and labs) and avoids capturing human faces. The participants have reviewed and signed the consent forms and information sheet so they are fully aware of the data capture process and future usage. The consent form that is approved by IRB and the information sheet can be provided upon request.

\subsection{Data Capture Platform}
In Table~\ref{tab:streams}, we provide additional information about the various streams available in the HoloAssist dataset. 

\minisection{Spatial coordinate systems}. The data capture platform utilizes a spatial anchor approach to establish a world coordinate system. All sensor modalities that are spatial in nature (head pose, gaze direction, hand poses, camera poses) are expressed with respect to the world coordinate system. The basis vector interpretation for coordinate system axes follows the convention of X=Forward, Y=Left, and Z=Up.

\minisection{Camera intrinsics} are represented with several parameters, including a 3×3 intrinsics matrix that converts camera coordinates (in the camera's local space) into normalized device coordinates (NDC) ranging from -1 to +1, radial and tangential distortion parameters, focal length, the principal point, and the image width and height.

\begin{longtable}{@{}p{0.15\linewidth}|p{0.2\linewidth}|p{0.25\linewidth}|p{0.25\linewidth}|p{0.1\linewidth}@{}}
\toprule
Sensor & Stream & Description & Representation & FPS \\
\midrule
\multirow{6}{*}{Color Camera} & \multirow{2}{*}{Color Image} & Image captured by front-facing HoloLens 2 camera & \multirow{2}{*}{RGB image $896 \times 504$ pixels} & \multirow{2}{*}{29.5 Hz} \\ \cmidrule{2-5}
 & \multirow{2}{*}{Color Camera Intrinsics} & Intrinsic parameters for front-facing HoloLens 2 camera & Camera intrinsics parameters (listed above) & \multirow{2}{*}{29.5 Hz} \\ \cmidrule{2-5}
 & \multirow{2}{*}{Color Camera Pose} & Spatial pose for color camera location &  \multirow{2}{*}{$4 \times 4$ coordinate system matrix} &  \multirow{2}{*}{29.5 Hz} \\
\midrule
\multirow{9}{*}{Depth Camera} & \multirow{5}{*}{Depth Image} & Depth image captured by HoloLens 2 depth camera in AHAT mode (provides pseudo-depth with phase wrap beyond 1 meter) &  \begin{tabular}[c]{@{}p{\linewidth}@{}}  \\16bpp Grayscale image $504 \times 504$ pixels \end{tabular} & \multirow{5}{*}{32.4 Hz} \\ \cmidrule{2-5}
& \multirow{2}{*}{Depth Camera Intrinsics} & Intrinsic parameters for HoloLens 2 depth camera & Camera intrinsics parameters (listed above) & \multirow{2}{*}{32.4 Hz} \\ \cmidrule{2-5}
& \multirow{2}{*}{Depth Camera Pose} & Spatial pose for depth camera location & $4 \times 4$ coordinate system matrix & \multirow{2}{*}{32.4 Hz} \\
\midrule
\multirow{2}{*}{Audio} & \multirow{2}{*}{Audio} & Audio stream captured by the microphone array & Single-channel, 48Khz, IEEE Float Wave Format & \multirow{2}{*}{48 KHz} \\
\midrule
\multirow{6}{*}{IMU} & \multirow{2}{*}{Accelerometer} & X-, Y-, and Z-axis inertial force in m/s$^2$ & \multirow{2}{*}{$3 \times 1$ vector} &\multirow{2}{*}{1.1 KHz} \\ \cmidrule{2-5}
& \multirow{2}{*}{Gyroscope} & X-, Y-, and Z-axis angular momentum in rad/s & \multirow{2}{*}{$3 \times 1$ vector} & \multirow{2}{*}{6.4 KHz} \\ \cmidrule{2-5}
& \multirow{2}{*}{Magnetometer} & X-, Y-, and Z-axis magnetic flux density in microteslas & \multirow{2}{*}{$3 \times 1$ vector} & \multirow{2}{*}{49.5 Hz} \\ 
\midrule
Head & Head Pose & Spatial pose for the user's head & $ 4 \times 4$ coordinate system matrix & 17.2 Hz \\
\midrule
\multirow{2}{*}{Eye Gaze}  & \multirow{2}{*}{Gaze Direction} & Spatial direction of the user's gaze & $3 \times 1$ vector for position and $ 3  \times 1$ vector for direction & \multirow{2}{*}{30.7 Hz} \\
\midrule
\multirow{3}{*}{Hands}  & \multirow{3}{*}{Hands Pose} & Spatial poses for the user's left and right-hand joints & $ 4 \times 4$ coordinate system for each of the 26 left-hand joints and 26 right-hand joints & \multirow{3}{*}{43.2 Hz}\\
\bottomrule
\caption{Details about the 7 sensor modalities captured live during data collection. The 8th modality, text, is added after the data collection. } \label{tab:streams}
\end{longtable}

\subsection{Annotation}

\begin{figure}[t]
    \centering
    \includegraphics[width=1.\linewidth]{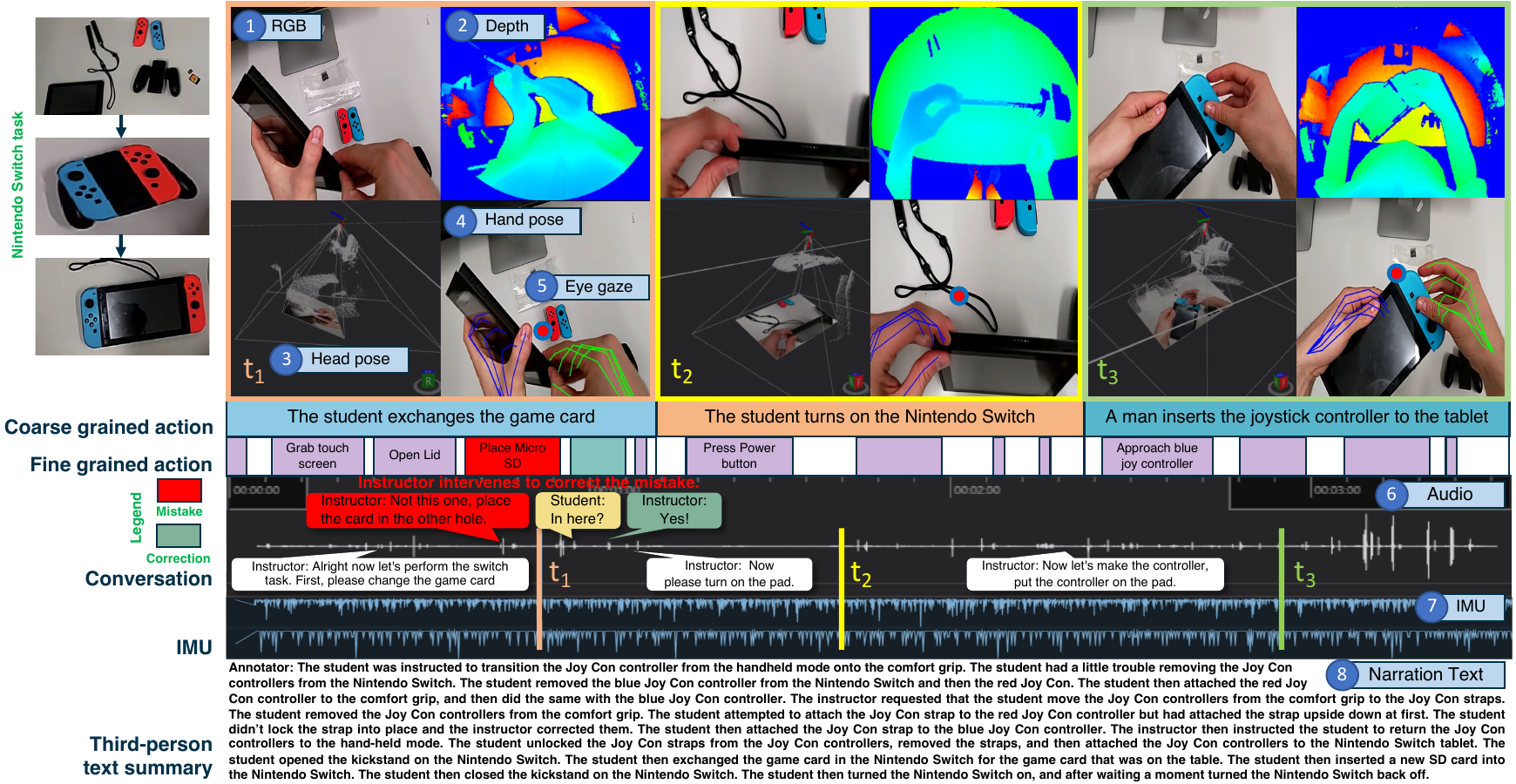}
    \caption{\data includes action and conversational annotations, in addition to text summaries of the videos, to indicate the mistakes and interventions in task completion. \emph{mistake} or \emph{correct} attributes are associated with each fine-grained action. A purpose label is associated with every conversation to indicate the type of verbal intervention. 
    }
    \label{fig:annotation_overview}
\end{figure}

\subsubsection{Annotation Definition}
Figure~\ref{fig:annotation_overview} provides an overview of the \data task and annotation structure. There are four event types in total for annotations: text summaries, coarse-grained Actions, fine-grained actions, and conversation. 

\minisection{Text summary.}
A summary of all activities performed by the task performer throughout the video. This annotation will describe all the actions across the video. The video annotator should use the Coarse-Grained Action annotations and the Conversation annotations to help build the Narration annotation. 

\minisection{Coarse-grained actions.}
A coarse-grained action is a step in the process that the task performer performs to help complete the task. Note that often the instructor will state the task before the task performer does the action–this is \emph{not} considered part of the Coarse-Grained Action.
Examples of coarse-grained actions include loading paper into a printer, screwing furniture legs into place, turning on a device, etc.

\minisection{Fine-grained actions.}
Fine-grained actions are the indivisible movements of the task performer during a coarse-grained action. These actions typically last for around 1-2 seconds or shorter, depending on the action. These will also include a designation of whether the action was correct or whether it was corrected later on by the task performer themselves or the instructor. 
Examples of fine-grained actions include the hand approaching button, hand pressing button, a task performer walking across the room, etc.

\minisection{Conversation.}
A Conversation annotation denotes the time frame when either the instructor or the task performer is speaking. The annotation specifies the content of the conversation.

\subsubsection{Annotation Structure}
The annotations follow the following format.

\minisection{Narration annotation structure.}
\begin{itemize}
    \item \textbf{Long-form description}: Use multiple sentences and make this as long as is necessary to be exhaustive. There are a finite number of scenarios across all videos, so make sure to call out the distinctive changes between videos, in particular, mistakes that the task performer makes in the learning process that are either self-corrected or corrected by the instructor.
    \item \textbf{Example}:
A man operates a big office printer. The instructor provides directions on how to turn on and load paper into the big office printer. The man turns on the printer and then turns it off. He then loads paper into the first drawer of the printer and replaces the first black cartridge from left to right in the printer. The instructor corrects the man on where to place the first black cartridge. The man moves the black cartridge from its current position to the correct location.
Note: The time stamps for this annotation will always start at 0 and end at the end of the video.
\end{itemize}

\minisection{Coarse-grained action annotation structure.}
\begin{itemize}
    \item \textbf{Coarse-Grained Action Sentences}: A factual statement describing the interaction that the collector/camera wearer is performing with a digital device and the software on the device. 
    \begin{itemize}
        \item Example 1: A man changes the battery of the bright green GoPro.
        \item  Example 2: A woman attaches the leg to a chair.
    \end{itemize}
    \item \textbf{Coarse-Grained Action Verb}: This verb was part of the Coarse-Grained Action sentence.
    \begin{itemize}
        \item Example 1: Change
        \item Example 2: Attach
    \end{itemize}
    \item \textbf{Coarse-Grained Action Adjective}: This is the adjective(s) that helps distinguish the noun from other similar items. This field is optional if the noun is unique enough on its own. 
    \begin{itemize}
        \item  Example 1: bright green
        \item Example 2: [blank]
    \end{itemize}
    \item \textbf{Coarse-Grained Action Noun}: This is the generic noun that is part of the Coarse-Grained Action sentence.
    \begin{itemize}
        \item Example 1: GoPro 
        \item Example 2: Leg
    \end{itemize}
\end{itemize}

\minisection{Fine-grained action annotation structure.}
\begin{itemize}
    \item \textbf{Fine-Grained Action Verb}: This is the verb that occurred during the Fine-Grained Action in the video.
    \begin{itemize}
        \item Example 1: Change
        \item Example 2: Attach
    \end{itemize}
    \item \textbf{Fine-Grained Action Adjective}: This is the adjective(s) that helps distinguish the noun from other similar items. This field is optional if the noun is unique enough on its own.
    \begin{itemize}
        \item Example 1: Bright Green
        \item Example 2: [blank]
    \end{itemize}
    \item \textbf{Fine-Grained Action Noun}: This is the generic noun that is part of the Fine-Grained Action in the video.
    \begin{itemize}
        \item Example 1: GoPro 
        \item Example 2: Leg
    \end{itemize}
    \item \textbf{Fine-Grained Action Attribute}: This attribute denotes whether the Fine-Grained action was the correct action and if it was the incorrect action, whether it was corrected, and by whom. You only need to select a single option from the list of options.
    \begin{itemize}
        \item Correct action
        \item Wrong action, corrected by instructor verbally
        \item Wrong action, corrected by performer
        \item Wrong action, not corrected
        \item Others
    \end{itemize}
\end{itemize}

\minisection{Conversation annotation structure.}
\begin{itemize}
    \item \textbf{Conversation Transcriptions}: Transcribe the conversation into texts.
    \item \textbf{Conversation Attribute}: Select an option that best describes the purpose of the speech. This is limited to the individual speaking and does not include any pause time waiting for a response.
    \item \textbf{Intervention Types}:
    \begin{itemize}
        \item Instructor-start-conversation:  Describing high-level instruction 
        \item Instructor-start-conversation:   Opening remarks
        \item Instructor-start-conversation:   Closing remarks
        \item Instructor-start-conversation:  Adjusting to capture better quality video
        \item Instructor-start-conversation:  Confirming the previous or future action
        \item Instructor-start-conversation:   Correct the wrong action
        \item Instructor-start-conversation:   Follow-up instruction
        \item Instructor-start-conversation: Other 
        \item Instructor-reply-to-task performer: Confirming the previous or future action
        \item Instructor-reply-to-task performer: Correct the wrong action
        \item Instructor-reply-to-task performer: Follow-up instruction
        \item Instructor-reply-to-task performer: other
        \item task performer-start-conversation: ask questions
        \item task performer-start-conversation: others
    \end{itemize}
\end{itemize}

\subsubsection{Label Statistics}
\minisection{Fine-grained action annotations.} In Figure~6 in the main paper, we present the distributions of the fine-grained actions.  We can see that the fine-grained actions follow a long-tail distribution. This is partly due to the open-world nature of the interaction and scenes.  Also, there are cases where the same action can be referred to with different expressions. For example, 'screw screw' and 'screw hex-cap-screw' are similar phrases. We asked the annotators to revisit the distributions of the nouns and verbs when the vocabulary is constructed and merge the ones referring to the same actions but with different expressions.  Still, this is a natural outcome of human annotations, and future research on addressing the label confusion issue is needed.

\minisection{Coarse-grained action annotations.} The coarse-grained actions are usually defined as high-level steps in the task, usually lasting around 30 seconds.  In Figure~7 in the main paper, we show the distributions of the coarse-grained actions and verbs and nouns of the actions. We can see that the coarse-grained actions follow a long-tail distribution similar to the fine-grained actions.  

\label{sec:appendix}

\minisection{Examples of conversation transcriptions and third-person text summary.}
Here we provide examples of the text summaries and the conversation transcriptions.  We can see that the third-person
text summaries capture the key events in the video, while the conversation transcripts are more interactive. \data 
provides both materials and can be useful in different contexts and applications. 
\begin{longtable}{|p{0.45\linewidth}|p{0.55\linewidth}|}
\hline
Third-person text summary & Conversation transcription \\ \hline
\begin{tabular}[c]{@{}l@{}}A task performer operated a capsule coffee machine. \\ The task performer grabbed a glass of water, and \\ poured it into the water container. The task performer \\ grabbed a coffee capsule and inserted it into \\ the capsule slot. The task performer pulled the lever to \\ use the capsule. With this, the task performer made \\ a coffee cup. The instructor asked the task performer \\ to empty the capsule container. The task performer \\ withdrew the drip tray,  then, removed the \\ capsule. And then inserted the drip tray back.\end{tabular} & \begin{tabular}[c]{@{}l@{}}task performer:  That one \\ task performer:  Start already \\ Instructor:  Add water\\ Instructor:  Grab a capsule and make a cup of coffee\\ Instructor:  Right\\ Instructor:  now, empty the capsule tray\\ Instructor:  Throw out the coffee\\ Instructor:  Put it on the table\\ Instructor:  Put it back\\ Instructor:  Your task is done press stop.\end{tabular} \\ \hline
\begin{tabular}[c]{@{}l@{}}The task performer grabbed the water container and \\ took it to the sink to fill it with water. Then \\ the task performer grabbed a coffee capsule and a \\ cup. The task performer placed the cup on the drip \\ tray and inserted the capsule into the capsule \\ slot pushing it down. The task performer reinserted \\ the capsule container by lowering the drip \\ tray as well. Then the task performer reinserted the \\ coffee capsule into the capsule slot. The \\ task performer lowered the lever and placed the \\ cup on the grid. Then pressed the right \\ button to start preparing the cup of coffee. \\ Finally, the task performer removed the drip tray \\ and the capsule container from the espresso \\ machine and separated them. Then the \\ task performer placed the capsule container on \\ the table and carried the drip tray to the \\ sink to empty it. Later the task performer inserted \\ the capsule container back into the drip \\ tray and finally inserted them into the \\ espresso machine.\end{tabular} & \begin{tabular}[c]{@{}l@{}}Instructor:  Hello, and thank you for agreeing to take part in our study. \\ In front of you, you will see an espresso coffee machine we will \\ be using for this task. First please load water into the machine\\ Instructor:  Please as you put the water facing down lift the lid.  \\ Instructor:  Please lift the lid\\ Instructor:  Please lift the top and slide it down \\ Instructor:  Please lift the top as you slide it down.\\ Instructor:  Next, please make a cup of coffee\\ Instructor:  Please, please return the capsule, please \\ Instructor:  Please grab the capsule from the capsule area.\\ Instructor:  Please remove the cup\\ Instructor:  Please lift up words the door and to the, please \\ afterward please pull please lift it\\ Instructor:  Please lift the expansion,  and pull it from the machine\\ Instructor:  Please reinsert the capsule.\\ Instructor:  Please push down on the metal\\ Instructor:  push it towards me now\\ Instructor:  Please make sure the capsule is part of the\\ Instructor:  Please make sure to retrain the capsule, to drop the \\ capsule to the capsule area\\ Instructor:  Next, please empty the capsule tray\\ Instructor:  Lastly, please empty the drip tray in the sink\\ Instructor:  Please take it out of the machine as before\\ Instructor:  Please remove the plastic, the capsule tray \\ Instructor:  That concludes everything for this task, you can now \\ press the stop button\end{tabular} \\ \hline
\begin{tabular}[c]{@{}l@{}}The task performer began the task by moving \\ the parts of the Nintendo on the table \\ and then grabbing it to change the game \\ card for another. He then opened the \\ Nintendo kick stand and grabbed a micro-SD\\ card, which he inserted into the SD slot. \\ Then he closed the kickstand. The task performer, \\ following the instructions of the instructor, \\ opened the kickstand behind the Nintendo \\ and stood it on the table. The man turned \\ on the Nintendo by pressing the power button. \\ Then, following the instructions of the instructor, \\ he turned off the console by pressing the power \\ button for three seconds and pressing the touch \\ screen of the console. The man grabbed the \\ touch screen and tried to place the blue controller \\ on it, until the instructor corrected him. Then he \\ placed both controls on the sides of the touch \\ screen of the console. The task performer removed \\ the controllers from the touch screen and inserted \\ them on the comfort grip, one on each side, \\ as instructed by the instructor. Then he removed \\ the joy controllers from the comfort grip and \\ then, correcting the errors he had when placing \\ the blue joy with controller, he correctly placed \\ each of them on the joy with straps as the \\ final action of the task.\end{tabular} & \begin{tabular}[c]{@{}l@{}}task performer:  Capturing\\ Instructor:  Alright now let's perform the switch task. First, please \\ change the game card \\ task performer:  In here?\\ Instructor:  Yes\\ task performer:  Game card there you go. \\ task performer:  Done\\ Instructor:  Okay\\ Instructor:  Now please change the SD card.\\ task performer:  There is no SD card\\ Instructor:  Please insert one\\ Instructor:  Okay\\ Instructor:  Now please make the path to stand up.\\ Instructor:  Open the support where the SD card is.\\ Instructor:  And now please turn on the pad\\ Instructor:  And now please turn it off.\\ Instructor:  You need to press more than three seconds \\ to turn it off.\\ Instructor:  Alright\\ Instructor:  Now let's make the controller, put the \\ controller on the pad.\\ Instructor:  No, the blue one go left\\ task performer:  Done\\ Instructor:  Okay\\ Instructor:  Now let's change the controller to the joystick.\\ task performer:  Change the... so I should remove it from the pad?\\ Instructor:  Yes, remove it first.\\ task performer:  And then?\\ Instructor:  Put it down the joystick\\ task performer:  On this?\\ Instructor:  Joystick\\ task performer:  Yeah, this is the joystick, right?\\ task performer: It's done\\ Instructor:  Now change it to another stick\\ task performer:  So I should? okay.\\ Instructor:  Take it out first.\\ task performer:  Okay\\ Instructor:  Okay.\\ Instructor:  Now you can press stop.\end{tabular} \\ \hline
\end{longtable}

\subsection{Experiment Details}
\subsubsection{Implementation Details} 
\minisection{ViT and TimeSformer.} In our action recognition, anticipation, mistake detection, and intervention prediction benchmarks, we use the vanilla Vision Transformers~\cite{vit} and the TimeSformer~\cite{timesformer} with divided space-time
 attention as the base model architecture.  Specifically, we use the \texttt{vit$\_$base$\_$patch16$\_$224} defined in the TimeSformer \href{https://github.com/facebookresearch/TimeSformer}{codebase}. The models are trained for 15 epochs with a base learning rate of 0.01. The learning rate is divided by 10 at epochs 11 and 14. We sample 8 frames as inputs depending on the need of the benchmarks. For example, in the action recognition benchmark, we sample 8 frames 
from the input action clip. For action anticipation and intervention prediction, we sample the frames from N seconds before the current action clip.  For mistake detection, 
we sample 8 frames from the beginning of the sequence until the end of the current action clip.  The number of frames may affect the model performance, and we use 8 to fit our compute resources better. 

 \minisection{Seq2Seq model.} We use a Seq2Seq model for 3D hand forecasting. We followed the basic Seq2Seq model in the literature~\cite{sutskever2014sequence} with modifications. We train the Seq2Seq network using ADAM optimizer~\cite{kingma2014adam} with a learning rate of 0.1. We implement the model with 512 hidden dimensions and 3 LSTM layers. We set the output dimension of the decoder as 156, the same as the number of the hand joint dimension. For the objective function, we use an L2 distance loss. We uniformly sample 8 frames as inputs in 3 seconds before the fine-grained action starts and validate hand forecasting with 1.5 seconds of hand poses from the start of the fine-grained action. If the invalid flag comes out of the device or the hand location is further than 1.5 from the head position, we consider this as the wrong hand pose, and we do not evaluate them.
 \begin{itemize}
     \item  \textbf{Encoder.} All inputs passed MLP with the same dimension number as the inputs except for the RGB images before the LSTM layers. For RGB images, we use the pre-trained ResNet~\cite{he2016deep} with 1000 output dimensions to extract RGB features. Then we concatenate features from MLP and ResNet and feed them into the LSTM layers. 
     \item  \textbf{Decoder.} For inputs of the LSTM layers, we use the same MLP and the ResNet architecture as the one in the Encoder. Finally, We use the outputs of the LSTM layers as the input to an additional MLP to predict hand joints.
 \end{itemize}

\subsubsection{3D Hand Pose Forecasting Visualization}
In Figure~\ref{fig:3d-hand-guidance-qualitative}, we provide visualizations of the Seq2Seq model used in the main paper on the 3D hand pose forecasting task.  As we can see from the figure, the accuracy of the 
hand pose forecasting still has a lot of room for improvement. 

\begin{figure}[h]
    \centering
    \includegraphics[width=\linewidth]{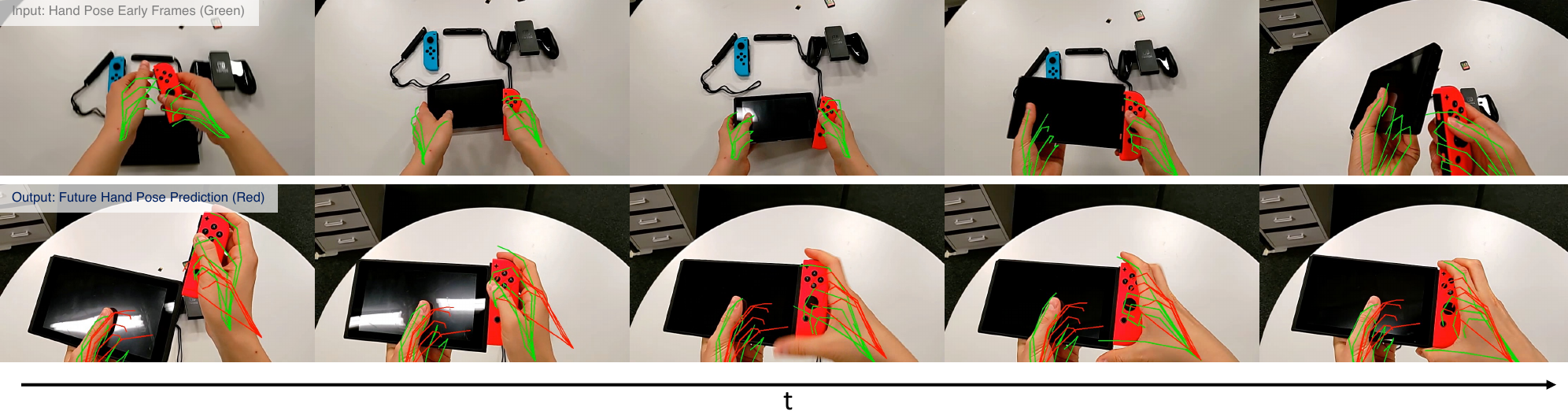}
    \caption{\textbf{3D hand pose forecasting qualitative results}. The input to the model is the historical ground-truth hand joint positions (\ie, hand poses, visualized in \textcolor{green}{Green}) with RGB images in the multimodal setting (RGB images are optional). Then the output is the prediction of future hand poses visualized in \textcolor{red}{Red}.}
    \label{fig:3d-hand-guidance-qualitative}
\end{figure}

\subsection{Data Sample Visualization}
\begin{figure}[h]
    \centering
    \includegraphics[width=.95\linewidth]{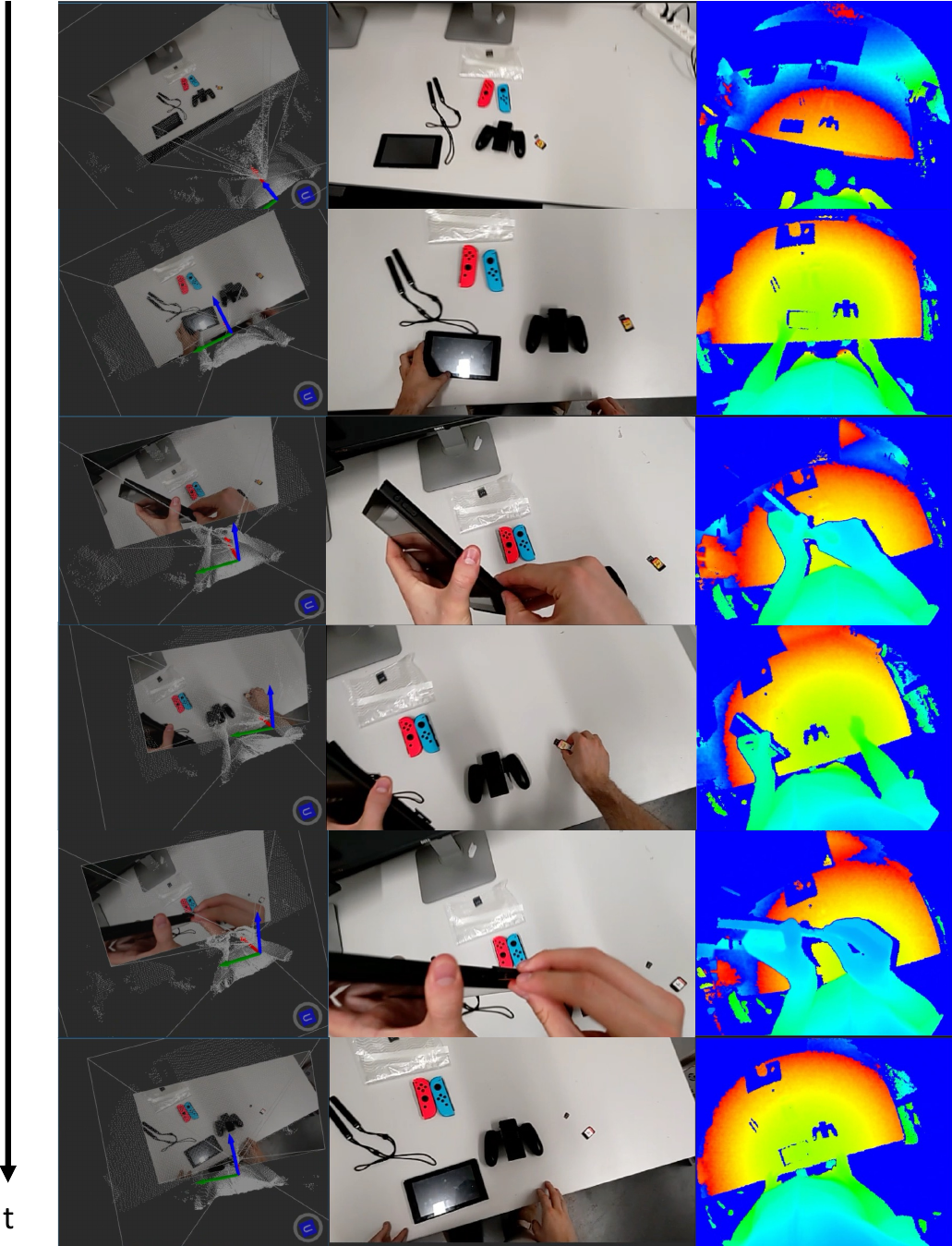}
    \caption{\textbf{Sample Data in \data}. The \textbf{left} column shows the point cloud and head poses. The \textbf{middle} column shows the RGB images. The \textbf{right} column shows the depth images.}
    \label{fig:sample_vis}
\end{figure}

In Figure~\ref{fig:sample_vis}, we provide a visualization of the session sequences in the dataset.  Here we visualize the head pose, point cloud, RGB images, and depth images in the figure.  The hand poses and eye gazes are omitted here, and you can refer to the figures in the main paper for reference. 



\end{document}